\documentclass{article}

\usepackage{microtype}
\usepackage{graphicx}
\usepackage{subcaption}
\usepackage{booktabs} 

\usepackage{hyperref}

\usepackage[accepted]{icml2021}

\usepackage[utf8]{inputenc} 
\usepackage[T1]{fontenc}    
\usepackage{url}            
\usepackage{amsmath,amsthm,amsfonts,amssymb,bm,stmaryrd}       
\usepackage{nicefrac}       
\usepackage{enumitem}
\usepackage[noend]{algpseudocode}
\usepackage{mathtools}
\usepackage{nccmath}
\usepackage{multirow}
\usepackage{bigdelim}
\newcommand*\Funct[2]{\textsc{#1}(#2)}
\newcommand*\Let[2]{\State #1 $\gets$ #2}

\usepackage{xcolor}		
\definecolor{darkblue}{rgb}{0, 0, 0.5}
\hypersetup{colorlinks=true, citecolor=darkblue, linkcolor=darkblue, urlcolor=darkblue}
\usepackage{booktabs}
\usepackage{dcolumn}
\newcolumntype{d}{D{.}{.}{-1}}
\usepackage{graphicx}
\usepackage{titlesec}
\usepackage{hyperref}
\usepackage{url}
\newtheorem{theorem}{Theorem}
\numberwithin{theorem}{section}

\newtheorem{lemma}[theorem]{Lemma}

\newtheorem{proposition}[theorem]{Proposition}

\theoremstyle{definition}
\newtheorem{definition}[theorem]{Definition}
\theoremstyle{remark}
\newtheorem{remark}[theorem]{Remark}

\newcommand{\smeta}{\widehat{\mathcal{S}}}
\newcommand{\pmeta}{\widehat{\mathcal{P}}}
\newcommand{\ptask}{P_\mathcal{T}}
\newcommand{\itrain}{\mathcal{I}_{\mathrm{train}}}
\newcommand{\ical}{\mathcal{I}_{\mathrm{cal}}}
\newcommand{\cset}{\mathcal{C}_{\epsilon}}
\newcommand{\mset}{\mathcal{M}_{\epsilon}}

\newcommand{\vtest}{\widehat{V}^{\mathrm{test}}_{i,k+1}}
\newcommand{\newpar}[1]{\textbf{{#1}.~}}

\titlespacing*{\subsection}{0pt}{.15\baselineskip}{.15\baselineskip}
\titlespacing*{\subsubsection}{0pt}{.15\baselineskip}{.15\baselineskip}
\titlespacing*{\section}{0pt}{.15\baselineskip}{.15\baselineskip}

\definecolor{darkgreen}{rgb}{0.31, 0.47, 0.26}
\renewcommand\cite{\citep}
\usepackage{fontawesome}

\begin{document}

\twocolumn[
\icmltitle{Few-shot Conformal Prediction with Auxiliary Tasks}




\begin{icmlauthorlist}
\icmlauthor{Adam Fisch}{mit}
\icmlauthor{Tal Schuster}{mit}
\icmlauthor{Tommi Jaakkola}{mit}
\icmlauthor{Regina Barzilay}{mit}
\end{icmlauthorlist}

\icmlaffiliation{mit}{Computer Science and Artificial Intelligence Laboratory, Massachusetts Institute of Technology, Cambridge, MA, USA}

\icmlcorrespondingauthor{Adam Fisch}{\texttt{fisch@csail.mit.edu}}

\icmlkeywords{Machine Learning, conformal prediction, ICML}

\vskip 0.3in
]



\printAffiliationsAndNotice{}  

\begin{abstract}

We develop a novel approach to conformal prediction when the target task has limited data available for training. Conformal prediction identifies a small set of promising output candidates in place of a single prediction, with guarantees that the set contains the correct answer with high probability. When training data is limited, however, the predicted set can easily become unusably large. In this work, we obtain substantially tighter prediction sets while maintaining desirable marginal guarantees by casting conformal prediction as a meta-learning paradigm over exchangeable collections of auxiliary tasks. Our conformalization algorithm is simple, fast, and agnostic to the choice of underlying model, learning algorithm, or dataset. We demonstrate the effectiveness of this approach across a number of few-shot classification and regression tasks in natural language processing, computer vision, and computational chemistry for drug discovery.

\end{abstract}

\section{Introduction}
\begin{figure*}[!t]
    \centering
    \small
    \includegraphics[width=.9\linewidth]{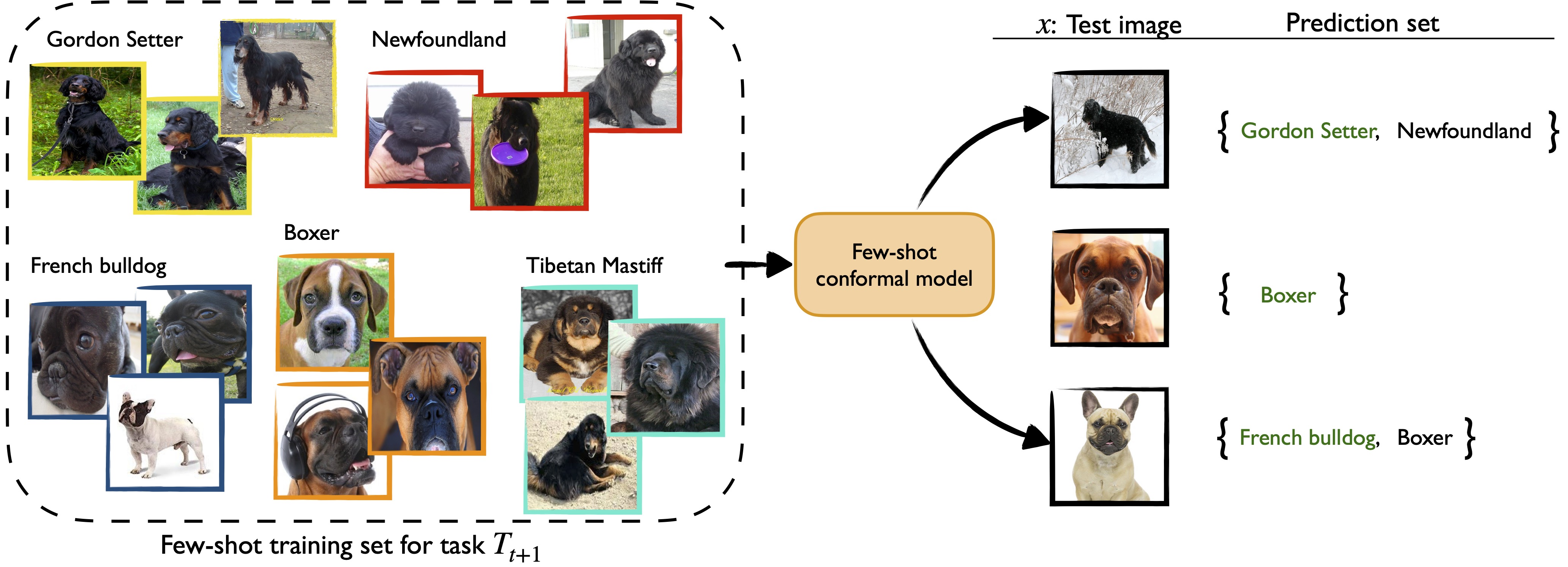}
    \vspace{-5pt}
    \caption{A demonstration of our conformalized few-shot learning procedure. Given a base model (e.g., a prototypical network for classification tasks~\cite{snell2017prototypical}) and a few demonstrations of a new task, our method produces a prediction \emph{set} that carries desirable guarantees that it contains the correct answer with high probability. Like other meta-learning algorithms, our approach leverages information gained from $t$ other, similar tasks---here to make more precise and confident predictions on the new task, $T_{t+1}$.}
    \vspace{-9pt}
    \label{fig:method}
\end{figure*}

\label{sec:introduction}
Accurate estimates of uncertainty are important for difficult or sensitive prediction problems that have variable accuracy~\cite[][]{ amodei201concrete, jiang2012medicine,jiang2018trust, angelopoulos2021sets}. Few-shot learning problems, in which training data for the target task is severely limited, pose a discouragingly compounded challenge: in general, not only is (1) making accurate predictions with little data hard, but also (2) rigorously quantifying the uncertainty in these few-shot predictions is even harder.

In this paper, we are interested in creating confident \emph{prediction sets} that provably contain the correct answer with high probability (e.g., 95\%), while only relying on a few in-task examples. Specifically, we focus on conformal prediction (CP)---a model-agnostic and distribution-free methodology for creating confidence-based set predictions~\cite{vovk2005algorithmic}. Concretely, suppose we have been given $n$ examples, $(X_j, Y_j) \in \mathcal{X} \times \mathcal{Y}$, $j=1,\ldots,n$, as training data, that have been drawn exchangeably from some underlying distribution $P$. Let $X_{n+1} \in \mathcal{X}$ be a new exchangeable test example for which we would like to predict  $Y_{n+1} \in \mathcal{Y}$.
The aim of conformal prediction is to construct a set-valued output, $\cset(X_{n+1})$, that contains  $Y_{n+1}$ with \emph{distribution-free marginal coverage} at a significance level $\epsilon \in (0, 1)$, i.e.,
\begin{equation}
    \label{eq:marginalcoverage}
    \mathbb{P}\left(Y_{n+1} \in \cset(X_{n+1})\right) \geq 1 - \epsilon.
\end{equation}
A conformal model is considered to be \emph{valid} if the frequency of error, $Y_{n+1} \not\in \cset(X_{n+1})$, does not exceed $\epsilon$. 
%
The challenge for few-shot learning, however, is that as $n \rightarrow 0$, standard CP methods quickly result in outputs $\cset(X_{n+1})$ so large that they lose all utility (e.g., a trivially valid classifier that returns all of $\mathcal{Y}$). A conformal model is only considered to be \emph{efficient} if $\mathbb{E}[|\cset(X_{n+1})|]$ is relatively small. 

%

In this work, we approach this frustrating data sparsity issue by casting conformal prediction as a meta-learning paradigm over exchangeable collections of tasks. 
By being exposed to a set of similar, auxiliary tasks, our model can \emph{learn to learn quickly} on the target task at hand. As a result, we can increase the data efficiency of our procedure, and are able to produce more precise---and confident---outputs.

Specifically, we use the auxiliary tasks to meta-learn both a \emph{few-shot model} and a \emph{quantile predictor}. The few-shot model provides relevance scores (i.e., \emph{nonconformity scores}, see \S\ref{sec:background_nonconf}) for each possible label candidate $y \in \mathcal{Y}$, and the quantile predictor provides a threshold rule for including the candidate $y$ in the prediction set, $\cset(X_{n+1})$, or not. A good few-shot model should provide scores that clearly separate \emph{correct} labels from \emph{incorrect} labels---much like a maximum-margin model. Meanwhile, a good quantile predictor---which is intrinsically linked to the specific few-shot model used---should quantify what few-shot scores correspond to relatively ``high'' or relatively ``low'' values for that task (i.e., as the name suggests, they infer the target quantile of the expected distribution of few-shot scores). Both of these models must be able to operate effectively given only a few examples from the target task, hence how they are meta-learned over auxiliary tasks becomes crucial.

Consider the example of image classification for novel categories (see Figure~\ref{fig:method} for an illustration). The goal is to predict the class of a new test image out of several never-before-seen categories---while only given a handful of training examples per category. In terms of auxiliary tasks, we  are given access to similarly-framed image classification tasks (e.g., \emph{cat} classes instead of \emph{dog} classes as in Figure~\ref{fig:method}). In this case, we can compute relevance by using a prototypical network~\cite{snell2017prototypical} to measure the Euclidean distance between the test image's representation and the average representation of the considered candidate class's support images (i.e., prototype).
Our quantile predictor then computes a ``distance cut-off'' that represents the largest distance between a label prototype and the test example that just covers the desired percentage of correct labels. Informally, on the auxiliary tasks, the prototypical network will learn efficient features, while the quantile predictor will learn what typically constitutes expected prototypical distances for correct labels when using the trained network.

We demonstrate that these two meta-learned components combine to make an efficient and simple-yet-effective approach to few-shot conformal prediction, all while retaining desirable theoretical performance guarantees. We empirically validate our approach on image classification, relation classification for textual entities, and chemical property prediction  for drug discovery. Our code is publicly available.\footnote{\url{https://github.com/ajfisch/few-shot-cp}.}

In summary, our main contributions are as follows:
\begin{itemize}[leftmargin=*]
    \item A novel theoretical extension of conformal prediction to include few-shot prediction with auxiliary tasks
    \item A principled meta-learning framework for constructing confident set-valued classifiers for new target tasks
    \item A demonstration of the practical utility of our framework across a range of classification and regression tasks.
\end{itemize}
\vspace{-4pt}
\section{Related Work}
\newpar{Uncertainty estimation} In recent years, there has been a growing research interest in estimating uncertainty in model predictions. A large amount of work has been dedicated towards \emph{calibrating} the model posterior,  $p_\theta(\hat{y}_{n+1}|x_{n+1})$, such that the true accuracy, $y_{n+1} = \hat{y}_{n+1}$, 
is indeed equal to the estimated probability~\cite[][]{niculescu2005predicting, lakshinarayanan2017ensemble, lee2018training}. In theory, these estimates could be used to create confident prediction sets $\cset(X_{n+1})$. Unlike CP, however, these methods are not guaranteed to be accurate, and often suffer from miscalibration in practice---and this is especially true for modern neural networks~\cite{pmlr-v70-guo17a, ashukha2020pitfalls,  hirschfeld2020uncertainty}. In a similar vein, Bayesian formalisms underlie several popular approaches to quantifying predictive uncertainty via computing the posterior distribution over model parameters~\cite[][]{neal1996bayesian, graves2011vi, hernandez2015bnn, gal2016dropout}. The quality of these methods, however, largely hinges on both (1) the degree of approximation required in computing the posterior, and (2) the suitability, or ``correctness'', of the presumed prior distribution. 

 \newpar{Conformal prediction} As introduced in \S\ref{sec:introduction}, conformal prediction~\cite{vovk2005algorithmic} provides a model-agnostic and finite-sample, distribution-free method for obtaining prediction sets with marginal coverage guarantees. Most pertinent to our work, \citet{linusson2014unstable} carefully analyze the effects of calibration set size on CP performance. For precise prediction sets, they recommend using at least a few hundred examples for calibration---much larger than the few-shot settings considered here. When the amount of available data is severely restricted, the predicted sets typically become unusably large. \citet{johansson2015small} and \citet{carlsson2015modifications} introduce similarly motivated approximations to CP with small calibration sets via interpolating calibration instances or using modified $p$-value definitions, respectively. Both methods are heuristics, however, and fail to provide finite-sample guarantees. Our work also complements  several recent directions that explore conformal prediction in the context of various validity conditions, such as conditional, risk-controlling, admissible, or equalized coverage~\cite[][\emph{inter alia}]{chernozhukov2019distributional, cauchois2020knowing, pmlr-v108-kivaranovic20a, romano2019quantile, Romano2020With, bates-rcps, fisch2021admission}.

\newpar{Few-shot learning} Despite the many successes of machine learning models, learning from limited data is still a significant challenge~\cite{bottou2007tradeoffs, lake2015omni, wang2020survey}. Our work builds upon the extensive few-shot learning literature by introducing a principled way of obtaining confidence intervals via meta-learning. Meta-learning has become a popular approach to transferring knowledge gained from auxiliary tasks---e.g., via featurizations or statistics~\cite{edwards2017statistician}---to a target task that is otherwise resource-limited~\cite[][]{vinyals2016matching, finn2017maml, snell2017prototypical, bertinetto2018metalearning, bao2020fewshot}. We leverage the developments in this area for our models (see Appendix~\ref{app:meta_algs}).

\section{Background}
\label{sec:background}

We begin with a review of conformal prediction~\cite[see][]{shafer2008tutorial}. 
Here, and in the rest of the paper, upper-case letters ($X$) denote random variables; lower-case letters ($x$) denote scalars, and script letters ($\mathcal{X}$) denote sets, unless otherwise specified. 
A list of notation definitions is given in Table~\ref{tab:symbols}.
All proofs are deferred to Appendix~\ref{app:proofs}. 


\subsection{Nonconformity measures}
\label{sec:background_nonconf}
Given a new example $x$, for every candidate label $y \in \mathcal{Y}$, conformal prediction applies a simple test to either accept or reject the null hypothesis that the pairing $(x, y)$ is correct. The test statistic for this hypothesis test is a \emph{nonconformity measure}, $\mathcal{S}\left((x, y), \mathcal{D}\right)$, where $\mathcal{D}$ is a dataset of exchangeable, correctly labeled examples. Informally, a lower value of $\mathcal{S}$ reflects that point $(x, y)$ ``conforms'' to $\mathcal{D}$, whereas a higher value of $\mathcal{S}$ reflects that $(x, y)$ is atypical relative to $\mathcal{D}$. A practical choice for $\mathcal{S}$ is model-based likelihood, e.g.,  $-\log p_\theta(y | x)$, where $\theta$ is a model fit to $\mathcal{D}$ using some learning algorithm $\mathcal{A}$ (such as gradient descent).
It is also important that $\mathcal{S}$ preserves exchangeability of its inputs. Let $Z_j := (X_j, Y_j)$, $j = 1,\ldots, n$ be the training data. Then, for test point $x \in \mathcal{X}$ and candidate label $y \in \mathcal{Y}$,  we calculate the \emph{nonconformity scores} for $(x, y)$ as:
\begin{align}
\label{eq:nonconf}
\begin{split}
    V^{(x, y)}_j &:= \mathcal{S}(Z_j, Z_{1:n} \cup \{(x, y)\}),\\
    V^{(x, y)}_{n+1} &:= \mathcal{S}((x, y), Z_{1:n} \cup \{(x, y)\}).
\end{split}
\end{align}
Note that this formulation, referred to as \emph{full} conformal prediction, requires running the learning algorithm $\mathcal{A}$ that underlies $\mathcal{S}$ potentially many times for every new test point (i.e., $|\mathcal{Y}|$ times). ``Split'' conformal prediction~\cite{Papadopoulos08}---which uses a held-out training set to learn $\mathcal{S}$, and therefore also preserves exchangeability---is a more computationally attractive alternative, but comes at the expense of predictive efficiency when data is limited.\footnote{\label{split_regression} Split conformal prediction also allows for simple nonconformity score calculations for regression tasks. For example, assume that a training set has been used to train a fixed regression model, $f_\theta(x)$. The absolute error nonconformity measure, $|y - f_\theta(x)|$, can then be easily evaluated for all $y \in \mathbb{R}$. Furthermore, as the absolute error monotonically increases away from $f_\theta(x)$, the conformal prediction $\cset$ simplifies to a closed-form \emph{interval}.}


\subsection{Conformal prediction}
\label{sec:background_cp}
To construct the final \emph{prediction} for the new test point $x$, the classifier tests the nonconformity score for each label $y$, $V_{n+1}^{(x, y)}$, against a desired significance level $\epsilon$, and includes all $y$ for which the null hypothesis---that the candidate pair $(x,y)$ is \emph{conformal}---is not rejected. This is achieved by comparing the nonconformity score of the test candidate to the scores computed over the first $n$ labeled examples.
This comparison leverages the quantile function, where for a random variable $V$ sampled from distribution $F$ we define
\begin{equation}
    \mathrm{Quantile}(\beta; F) := \inf\{v \colon F(v) \geq \beta\}.
\end{equation}
In our case, $F$ is the distribution over the $n+1$ nonconformity scores, denoted $V_{1:n+1}$. However, as we do not know  $V_{n+1}^{(x, y)}$ for the true $y^*$, we use an ``inflated'' quantile:
\begin{lemma}[Inflated quantile]
\label{lemma:quantile}
Assume that $V_j$, $j=1,$ $\ldots, n+1$ are exchangeable random variables.
Then for any $\beta \in (0, 1)$,  $\mathbb{P}\left(V_{n+1} \leq \mathrm{Quantile}(\beta, V_{1:n} \cup \{\infty\})\right) \geq \beta.$
\end{lemma}

Conformal prediction then guarantees marginal coverage by including all labels $y$ for which $V_{n+1}^{(x, y)}$ is below the inflated quantile of the $n$ training points, as summarized:

\begin{theorem}[CP, \citet{vovk2005algorithmic}]
\label{thm:conformalprediction}
Assume that examples $(X_j, Y_j)$, $j=1,\ldots, n+1$ are exchangeable. For any nonconformity measure $\mathcal{S}$ and $\epsilon \in (0, 1)$, define the conformal set (based on the first $n$ examples) at  $x \in \mathcal{X}$ as
\begin{equation*}
\label{eq:csetconstruction}
\resizebox{1\hsize}{!}{$\displaystyle
	\cset(x) := \Big \{ y \in \mathcal{Y} \colon 
    V_{n+1}^{(x,y)} \leq \mathrm{Quantile}(1 - \epsilon;\, V_{1:n}^{(x, y)} \cup \{\infty\}) \Big\}.
$}
\end{equation*}
Then $\cset(X_{n+1})$ satisfies Eq.~\eqref{eq:marginalcoverage}.
\end{theorem}

Though Theorem~\ref{thm:conformalprediction} provides  guarantees for any training set size $n$, in practice $n$ must be fairly large (e.g., 1000) to achieve reasonable, stable performance---in the sense that $\cset$ will not be too large on average~\cite{lei2018distribution,bates-rcps}. This is a key hurdle for few-shot conformal prediction, where $n = k$ is assumed to be small (e.g., 16).
\section{\mbox{Few-shot Meta Conformal Prediction}}
We now propose a general meta-learning paradigm for training efficient conformal predictors, while relying only on a very limited number of \emph{in-task} examples.

At a high level, like other meta-learning algorithms, our approach leverages information gained from $t$ other, similar tasks in order to perform better on task $t+1$. In our setting we achieve this by learning a more statistically powerful nonconformity measure and quantile estimator than would otherwise be possible using only the limited data available for the target task. Our method uses the following recipe:
\begin{enumerate}[leftmargin=*]
    \item We meta-learn (and calibrate) a nonconformity measure and quantile predictor over a set of auxiliary tasks;
    \item We adapt our meta nonconformity measure and quantile predictor using the examples we have for our target task;
    \item We compute a conformal prediction set for a new input $x \in \mathcal{X}$ by including all labels $y \in \mathcal{Y}$ whose meta-learned nonconformity score is below the predicted $1 - \epsilon$ quantile.
\end{enumerate}
Pseudo-code for our meta CP procedure is given in Algorithm~\ref{alg:meta}. This skeleton focuses on classification; regression follows similarly. 
Our framework is model agnostic, in that it
allows for practically any meta-learning implementation for both nonconformity and quantile prediction models. 

In the following sections, we break down our approach in detail.
In \S\ref{sec:task_formulation} we precisely formulate our few-shot learning setup with auxiliary tasks. In \S\ref{sec:metalearning} and \S\ref{sec:metacalibration} we describe our meta-learning and meta-calibration setups, respectively. Finally, in \S\ref{sec:extensions} we discuss further theoretical extensions.
For a complete technical description of our modeling choices and training strategy for  our experiments, see Appendix~\ref{app:meta_cp_extra}.

\subsection{Task formulation}
\label{sec:task_formulation}
In this work, we assume access to $t$ auxiliary tasks, $T_i$, $i = 1,\ldots,t$, that we wish to leverage to produce tighter uncertainty sets for predictions on a new task, $T_{t+1}$. Furthermore, we assume that these $t+1$ tasks are \emph{exchangeable} with respect to some task distribution, $\ptask$. Here, we treat $\ptask$ as a distribution over random  distributions, where each task $T_i \in \mathcal{T}$ defines a task-specific distribution, $P_{XY} \sim \ptask$, over examples $(X, Y) \in \mathcal{X} \times \mathcal{Y}$. The randomness is in both the task's relation between $X$ and $Y,$ and the task's data.

For each of the $t$ auxiliary tasks, we do not make any assumptions on the amount of data we have (though, in general, we expect them to be relatively unrestricted). On the new task $T_{t+1}$, however, we only assume a total of $k$ exchangeable training examples. Our goal is then to develop a task-agnostic uncertainty estimation strategy that generalizes well to new examples from the task's unseen test set, $(X^{\mathrm{test}}_{t+1}, Y^{\mathrm{test}}_{t+1})$.\footnote{For ease of notation, we write $X_{t+1}^{\mathrm{test}}$ to denote the $(k+1)$th example of task $T_{t+1}$, i.e., the new test point after observing $k$  training points. This is equivalent to test point $X_{n+1}$ from \S\ref{sec:background}.} Specifically, we desire  finite-sample marginal \emph{task} coverage, as follows:
\begin{definition}[Task validity] Let $\mset$ be a set-valued predictor. $\mset$ is considered to be \emph{valid} across tasks if for any task distribution $\ptask$ and $\epsilon \in (0, 1)$, we have
\begin{equation}
    \label{eq:task_valid}
    \mathbb{P}\Big(Y^{\mathrm{test}}_{t+1} \in \mset\left(X^{\mathrm{test}}_{t+1}\right) \Big) \geq 1 - \epsilon.
\end{equation}
\end{definition}
Note that we require the marginal coverage guarantee above to hold {on average} across tasks and their examples. 

\begin{figure}[t!]
    \centering
    \hypersetup{hidelinks}
    \begin{minipage}{1\linewidth}
    \begin{algorithm}[H]
    \footnotesize
    \caption{ \small Meta conformal prediction with auxiliary tasks.}
    \label{alg:meta}
    \textbf{Definitions:} $T_{1:t+1}$ are exchangeable tasks. $\itrain \cup \ical$ are the $t$ tasks used for meta-training and meta-calibration. $z_{1:k} \in (\mathcal{X} \times \mathcal{Y})^k$ are the $k$ support examples for target task $T_{t+1}$. $x \in \mathcal{X}$ is the target task input. $\mathcal{Y}$ is the label space. $\epsilon$ is the significance.
    \vspace{5pt}
    \begin{algorithmic}[1]

        \Function{predict}{$x$, $z_{1:k}$, $T_{1:t}$, $\epsilon$}
            \State \textcolor{darkgreen}{\emph{\# Learn $\smeta$ and $\pmeta$ on meta-training tasks (\S\ref{sec:metalearning}).}}
            \State \textcolor{darkgreen}{\emph{\# $\smeta$ and $\pmeta$ are meta nonconformity/quantile models.}}
            \Let{$\smeta$, $\pmeta_{1-\epsilon}$}{$\Funct{train}{T_i, i \in \itrain}$}
            \State \textcolor{darkgreen}{\emph{\# Predict the $1-\epsilon$ quantile.}}
            \Let{$\widehat{Q}_{t+1}$}{$\pmeta_{1 - \epsilon}(z_{1:k}; \phi_{\mathrm{meta}})$}
            \State \textcolor{darkgreen}{\emph{\# Initialize empty output set.}}
            \Let{$\mset$}{$\{\}$}
            \State \textcolor{darkgreen}{\emph{\# (Note that for regression tasks, where $|\mathcal{Y}| = \infty$,  for}}
            \State \textcolor{darkgreen}{\emph{\# certain $\smeta$ the following simplifies to a closed-form }}
            \State \textcolor{darkgreen}{\emph{\# interval, making it tractable---see \S\ref{sec:background_nonconf}, footnote~\ref{split_regression}.)}}
            \For{$y \in \mathcal{Y}$}
              \State \textcolor{darkgreen}{\emph{\# Compute the nonconformity score for label $y$.}}
              \Let{$\widehat{V}_{t+1, k+1}^{(x, y)}$}{$\smeta((x, y), z_{1:k}; \theta_{\mathrm{meta}})$}
                \State \textcolor{darkgreen}{\emph{\# Compare to the \underline{calibrated} quantile (\S\ref{sec:metacalibration}).}}
                \If{$\widehat{V}_{t+1, k+1}^{(x,y)} \leq \widehat{Q}_{t+1} + \Lambda(1 - \epsilon, \ical)$}
                        \Let{$\mset$}{$\mset \cup \{y\}$}
                    \EndIf
            \EndFor
            \State \Return{$\mset$}
        \EndFunction
    \end{algorithmic}
    \end{algorithm}
    \end{minipage}
\end{figure}

\subsection{Meta-learning conformal prediction models}
\label{sec:metalearning}

Given our collection of auxiliary tasks, we would like to meta-learn both (1) an effective nonconformity measure that is able to adapt quickly to a new task using only $k$ examples; and (2) a quantile predictor that is able to robustly identify the $1 - \epsilon$ quantile of that same meta nonconformity measure, while only using the same $k$ examples.


Prior to running our meta-learning algorithm of choice, we split our set of $t$ auxiliary tasks into disjoint sets of training tasks, $\itrain$, and calibration tasks, $\ical$, where $|\itrain| + |\ical| = t$. See Table~\ref{tab:data_splits} for an overview of the different splits. We use $\itrain$ to learn our meta nonconformity measures and quantile predictors, which we discuss now. 
Additional technical details are contained in Appendix~\ref{app:meta_algs}.

\begin{table}[!t]
\small
\centering
\begin{tabular}{lccc}
\addlinespace[-\aboverulesep] 
\cmidrule[\heavyrulewidth]{2-4}
& Task Split  & \# Tasks    & \# Examples / Task \\
\cmidrule{2-4}
\hspace{-1em}\ldelim\{{2}{*}[{Auxiliary}]  & Meta-training    & $|\itrain|$ & $\gg k$         \\
& Meta-calibration & $|\ical|$   & $k + m_i$            \\
\cmidrule{2-4}
& Test        & 1           & $k$                  \\
\cmidrule[\headrulewidth]{2-4}
\end{tabular}

\caption{An overview of the data assumptions for a single test task ``episode''. We use $|\itrain| + |\ical| = t$ total auxiliary tasks to create more precise uncertainty estimates for the  $(t+1)$th test task. This is repeated for each test task (\S\ref{sec:exp-set}). $m_i \gg k$ is the number of extra examples per calibration task that are used to compute an empirical CDF when finding $\Lambda(\beta; \ical)$---it may vary per task.}
\vspace{-5pt}
\label{tab:data_splits}
\end{table}


\newpar{Meta nonconformity measure} Let $\smeta\left((x, y), \mathcal{D}; \theta_{\mathrm{meta}}\right)$ be a \emph{meta nonconformity measure}, where $\theta_{\mathrm{meta}}$ are meta parameters learned over the auxiliary tasks in $\itrain$. Since $\theta_{\mathrm{meta}}$ is fixed after the meta training period, $\smeta$ preserves exchangeability over new collections of exchangeable tasks (i.e., $\ical$) and task examples. Let $Z_{i,j} := (X_{i, j}, Y_{i,j})$, $j = 1, \ldots, k$ be the few-shot training data for a task $T_i$ (here $i$ is the task index, while $j$ is the example index). Given a new test point $x \in \mathcal{X}$ and candidate pairing $(x, y)$, the \emph{meta nonconformity scores} for $(x, y)$ are
\begin{align}
\label{eq:meta_nonconf}
\begin{split}
    \widehat{V}^{(x, y)}_{i, j} &:= \smeta(Z_{i, j}, Z_{i, 1:k} \cup \{(x, y)\}; \theta_{\mathrm{meta}}), \\
    \widehat{V}^{(x, y)}_{i, k+1} &:= \smeta((x, y), Z_{i, 1:k} \cup \{(x, y)\}; \theta_{\mathrm{meta}}).
\end{split}
\end{align}

\begin{figure}[!t]
    \centering
    \small
    \includegraphics[width=0.9\linewidth]{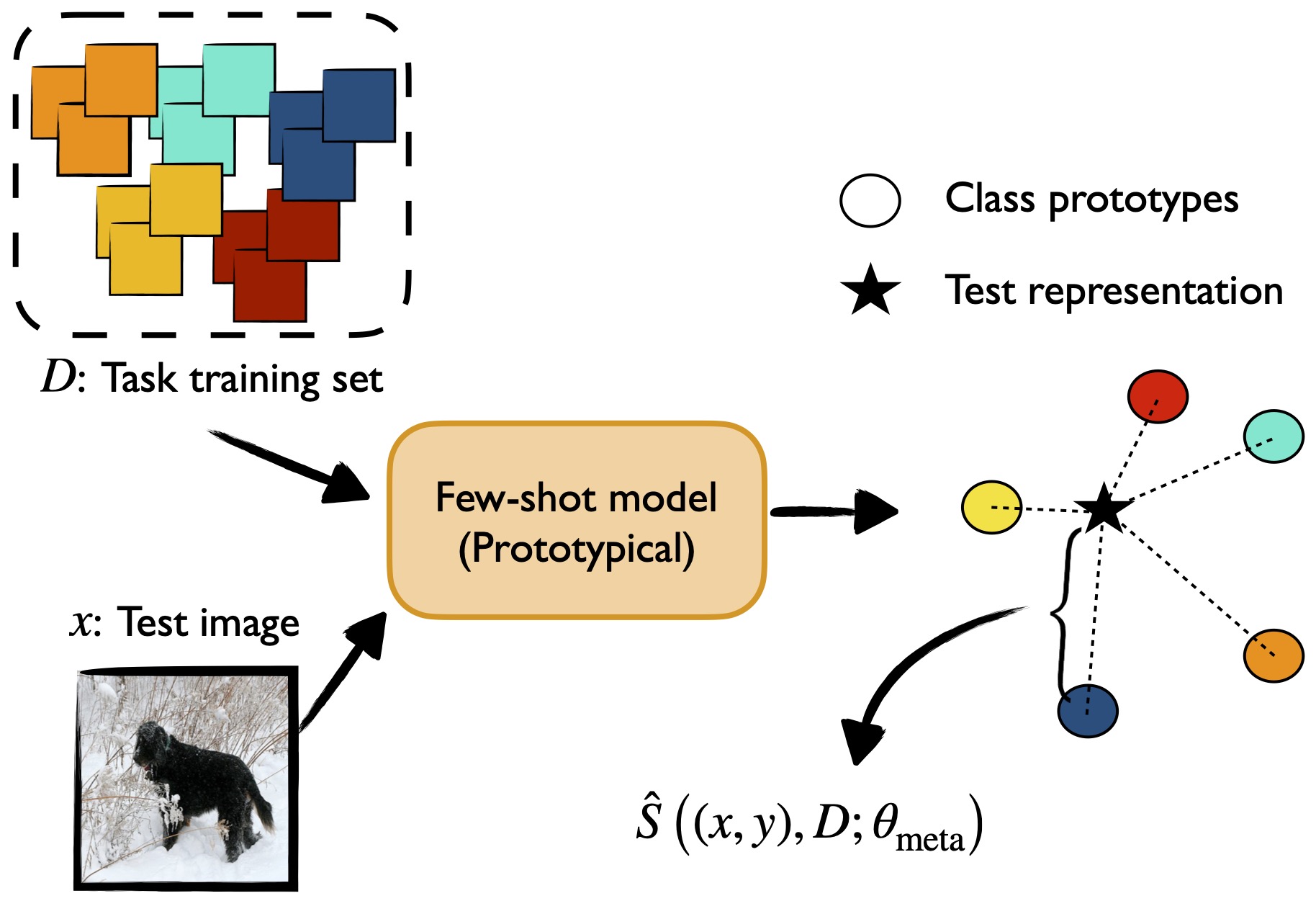}
    \vspace{-8pt}
    \caption{An example of using a prototypical network~\cite{snell2017prototypical} to compute meta nonconformity scores. If $\smeta$ is well-trained, the distance between the test point and the correct class prototype should be small, and the distance to incorrect prototypes large, even when the number of in-task training examples is limited.}
    \vspace{-10pt}
    \label{fig:scores}
\end{figure}

As an example, Figure~\ref{fig:scores} demonstrates how we compute $\widehat{V}_{i,k+1}^{(x,y)}$ using the distances from a meta-learned prototypical network following the setting in Figure~\ref{fig:method}.

Computing all $k+1$ scores $|\mathcal{Y}|$ times is typically tractable due to the few number of examples (e.g., $k \approx 16$) and the underlying properties of the meta-learning algorithm driving $\smeta$. For example, prototypical networks only require a forward pass. A naive approach to few-shot conformal prediction is to exploit this efficiency, and simply run full CP using all $k+1$ data points.  Nevertheless, though a strong baseline, using only $k+1$ points to compute an empirical quantile is still suboptimal. As we discuss next, instead we choose to regress the desired quantile directly from $Z_{i, 1:k}$, and disregard the empirical quantile completely. 

Since we predict the quantile instead of relying on the empirical quantile, we do not have to retain exchangeability for $Z_{i,1:k}$. As a result,  we switch to   ``split'' CP  (\S\ref{sec:background_nonconf}), and  \emph{do not} include $(x, y)$ when calculating ${\widehat{V}_{i,j}^{(x, y)}}$, as this is faster.


\newpar{Meta quantile predictor} Let $\pmeta_{\beta}(\mathcal{D}; \phi_{\mathrm{meta}})$ be a \emph{meta $\beta$-quantile predictor}, where $\phi_{\mathrm{meta}}$ are the meta parameters learned over the auxiliary tasks in $\itrain$. $\pmeta_{\beta}$ is trained to predict the $\beta$-quantile of $F$---where $F$ is the underlying task-specific distribution of nonconformity scores---given $\mathcal{D}$, a dataset of $Z = (X, Y)$ pairs sampled from that task.

As some intuition for this approach, recall that in calculating $\mathrm{Quantile}(\beta; F)$ given exchangeable samples $v_{1:n} \sim F$, we implicitly need to estimate $\mathbb{P}(V_{n+1} \leq v \mid v_{1:n})$. For an appropriate parametrization $\psi$ of $F$, de Finetti's theorem for exchangeable sequences allows us to write
\begin{equation*}
\resizebox{1\hsize}{!}{$\displaystyle
    \mathbb{P}(V_{n+1} \leq v \mid v_{1:n}) \propto \hspace{-.15cm} \int\displaylimits_{-\infty}^{v} \hspace{-.2cm} \int\displaylimits_{\hspace{.2cm}\Psi} p(v \mid \psi) \prod_{i=1}^{n} p(v_i \mid \psi)p(\psi)d\psi dv.
    $}
\end{equation*}
In this sense, meta-learning over auxiliary task distributions may help us learn a better prior over latent parametrizations $\psi$---which in turn may help us better model the $\beta$-quantile than we could have, given only $k$ samples and nothing else.

We develop a simple approach to modeling and learning $\pmeta_\beta$.
Given the training examples $Z_{i,1:k}$, we use a deep sets model~\cite{zaheer2017sets} parameterized by $\phi_{\mathrm{meta}}$ to predict the $\beta$-quantile of $\vtest$, the random variable representing the nonconformity score of the test point, $Z_{i, k+1} := (X_{i,k+1}, Y_{i, k+1})$. We optimize $\phi_{\mathrm{meta}}$ as
\begin{equation}
  \min_{\phi} \hspace{-.1cm} \sum_{i \in \itrain} \hspace{-.2cm}\Big(\pmeta_{\beta}\big(Z_{i, 1:k}; \phi\big) - \mathrm{Quantile}\big(\beta; \vtest\big)\Big)^2,
\end{equation}
where we estimate the target, $\mathrm{Quantile}\big(\beta; \vtest\big)$, using $m \gg k$ extra examples sampled from the training task.

\begin{figure}[!t]
    \centering
    \small
    \includegraphics[width=1\linewidth]{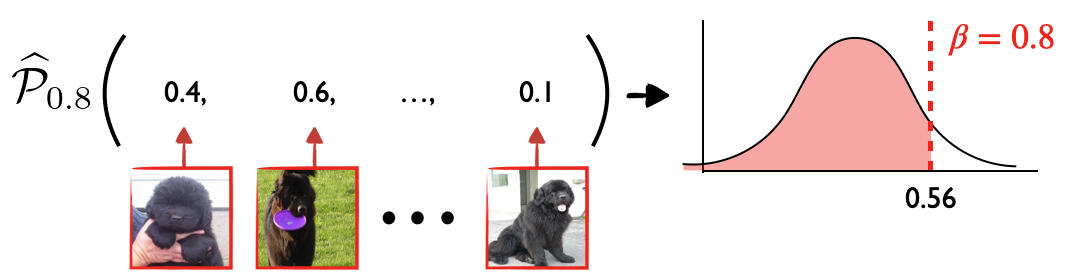}
    \caption{An illustration of using our meta-learned quantile predictor $\pmeta_{\beta}$ to infer the $\beta$-quantile of the distribution of $\vtest$, given the few examples from $T_i$'s training set. The numbers above each image reflect the leave-one-out scores we use as inputs, see Eq.~\eqref{eq:loo}.}
    \label{fig:quantile_infer}
    \vspace{-5pt}
\end{figure}

In practice, we found that choosing to first transform $Z_{i,1:k}$ to \emph{leave-one-out} meta nonconformity scores,
\begin{equation}
\label{eq:loo}
\widehat{L}_{i,j} := \smeta\big(Z_{i,j}, Z_{i, 1:k} \setminus Z_{i,j}; \theta_{\mathrm{meta}}\big),
\end{equation}
and providing $\pmeta_{\beta}$ with these scalar leave-one-out scores as inputs, performs reasonably well and is lightweight to implement. 
Inference using $\pmeta_{\beta}$ is illustrated in Figure~\ref{fig:quantile_infer}.


\newpar{Training strategy} The meta nonconformity measure $\smeta$ and meta quantile predictor $\pmeta_{\beta}$ are tightly coupled, as given a fixed $\smeta$, $\pmeta_{\beta}$ learns to model its behavior on new data. A straightforward, but data inefficient, approach to training $\smeta$ and $\pmeta_{\beta}$ is to split the collection of auxiliary tasks in $\itrain$ in two, i.e., $\itrain = \itrain^{(1)} \cup \itrain^{(2)}$, and then train $\smeta$ on $\itrain^{(1)}$, followed by training $\pmeta_{\beta}$ on $\smeta$'s predictions over $\itrain^{(2)}$. The downside of this strategy is that both $\smeta$ and $\pmeta_{\beta}$ may be sub-optimal, as neither can take advantage of all of $\itrain$.

We employ a slightly more involved, but more data efficient approach, where we split $\itrain$ into $k_f$ folds, i.e., $\itrain = \bigcup_{f=1}^{k_f} \itrain^{(f)}$. We then train $k_f$ separate meta nonconformity measures $\widehat{S}_f$, where we leave out fold $f$ from the training data. Using $\widehat{S}_f$, we compute nonconformity scores on fold $f$'s data, aggregate these nonconformity scores across all $k_f$ folds, and train the meta quantile predictor on this union. Finally, we train another nonconformity measure on \emph{all} of $\itrain$, which we use as our ultimate $\widehat{S}$. This way we are able to use all of $\itrain$ for training both $\smeta$ and $\pmeta_{\beta}$. This process is illustrated in Figure~\ref{fig:training}. Note that it is not problematic for $\pmeta_{\beta}$ to be trained on the collection of $\smeta$ instances trained on $k_f -1$ folds, but then later used to model one $\smeta$ trained on \emph{all} the data, since it will be calibrated (next, in \S\ref{sec:metacalibration}).

\subsection{Calibrating meta-learned conformal prediction}
\label{sec:metacalibration}
Though $\pmeta_{\beta}$ may obtain low empirical error after training, it 
does not have any inherent rigorous guarantees out-of-the-box. Given our held-out set of auxiliary tasks $\ical$, however, we can quantify the uncertainty in $\pmeta_{\beta}$ (i.e., how far off it may be from the true quantile), and calibrate it accordingly. The following lemma formalizes our meta calibration procedure:

\begin{lemma}[Meta calibration]
\label{lemma:meta_calibration}
Assume $\widehat{Q}_i$, $i \in \ical$ are the (exchangeable) meta $\beta$-quantile predictions produced by $\pmeta_{\beta}$ for tasks $T_i$, $i \in \ical$. Let $\vtest$ be the meta nonconformity score for a new sample from task $T_i$, where $F_i$ is its distribution function. Define the correction $\Lambda(\beta; \ical)$ as
\begin{equation}
\resizebox{.89\hsize}{!}{$\displaystyle
    \hspace{-8pt}\Lambda(\beta; \ical) := \inf\bigg\{\lambda \colon \frac{1}{|\ical|+1} \sum_{i \in \ical} F_i\big(\widehat{Q}_i + \lambda\big) \geq \beta\bigg\}.
$}
\end{equation}
We then have that $\mathbb{P}\big(\widehat{V}_{t+1,k+1}^{\mathrm{test}} \leq \widehat{Q}_{t+1} + \Lambda(\beta; \ical)\big) \geq \beta$.
\end{lemma}

It is important to pause to clarify at this point that calculating $\Lambda(\beta; \ical)$ requires knowledge of the true meta nonconformity distribution functions, $F_i$, for all calibration tasks. For simplicity, we write Lemma~\ref{lemma:meta_calibration} and the following Theorem~\ref{thm:meta_cp} as if these distribution functions are indeed known (again, only for calibration tasks). In practice, however, we typically only have access to an \emph{empirical} distribution function over $m_i$ task samples. In this case, Lemma~\ref{lemma:meta_calibration} holds in expectation over task samples  $Z_{i,k:k +m_i}$, as for an empirical distribution function of $m$ points, $\widehat{F}_m$, we have $\mathbb{E}[\widehat{F}_m] = F$. Furthermore, for large enough $m_i$, concentration results suggest that we can approximate $F_i$ with little error given a particular sample (this is the focus of $\S\ref{sec:extensions}$).


That said, in a nutshell, Lemma~\ref{lemma:meta_calibration} allows us to probabilistically adjust for the error in $\pmeta_{\beta}$, such that it is guaranteed to produce valid $\beta$-quantiles on average. We can then perform conformal inference on the target task by comparing each meta nonconformity score for a  point $x \in \mathcal{X}$ and candidate label $y \in \mathcal{Y}$ to the calibrated meta quantile, and keep all candidates with nonconformity scores that fall below it. 

\begin{theorem}[Meta CP]
\label{thm:meta_cp}
Assume that tasks $T_i$, $i \in \ical$ and $T_{t+1}$ are exchangeable, and that their nonconformity distribution functions $F_i$ are known. For any meta quantile predictor $\pmeta_{1-\epsilon}$, meta nonconformity measure $\smeta$, and $\epsilon \in (0, 1)$, define the meta conformal set (based on the tasks in $\ical$ and the $k$ training examples of task $T_{t+1}$) at $x \in \mathcal{X}$ as
\begin{equation*}
\resizebox{1\hsize}{!}{$\displaystyle
 \mset(x):= \Big \{ y \in \mathcal{Y} \colon \widehat{V}_{t+1,k+1}^{(x, y)} \leq \widehat{Q}_{t+1} + \Lambda(1 - \epsilon; \ical)\Big\},
$}
\end{equation*}
where $\widehat{Q}_{t+1}$ is the result of running $\pmeta_{1-\epsilon}$ on the $k$ training examples of task $T_{t+1}$. Then $\mset(X_{t+1}^{\mathrm{test}})$ satisfies Eq.~\eqref{eq:task_valid}.
\end{theorem}

It should be acknowledged that Theorem~\ref{thm:meta_cp} guarantees coverage \emph{marginally} over tasks, as specified in Eq.~\eqref{eq:task_valid}. Given appropriate assumptions on the quantile predictor $\pmeta_{1-\epsilon}$, we can achieve task-\emph{conditional} coverage asymptotically:
\begin{definition}[Consistency]
We say $\pmeta_{1- \epsilon}$ is an asymptotically consistent estimator of the $1-\epsilon$ quantile if
\begin{equation*}
    \big|\pmeta_{1- \epsilon}(Z_{i,1:k};\phi_{\mathrm{meta}}) -\mathrm{Quantile}(1 - \epsilon, F_i)\big| = o_{\mathbb{P}}(1)
\end{equation*}
as $k \rightarrow \infty$, where $F_i$ is the CDF of nonconformity scores for any task $t_i \in \mathcal{T}$. In other words, $\pmeta_{1- \epsilon}$ converges in probability to the true quantile given enough in-task data.
\end{definition}
\begin{proposition}[Asymptotic meta CP] 
\label{claim:asymptotic}
If $\pmeta_{1-\epsilon}$ is asymptotically consistent, then as $k\rightarrow \infty$ the meta conformal set $\mset$ achieves asymptotic conditional coverage, where
\begin{equation*}
\resizebox{1\hsize}{!}{$\displaystyle
  \hspace{-1pt}\mathbf{1}\Big\{\mathbb{P}\big(Y^{\mathrm{test}}_{t+1} \in \mset\left(X^{\mathrm{test}}_{t+1}\right) \mid T_{t+1} = t_{t+1}\big) \geq 1 - \epsilon\Big\} = 1 - o_{\mathbb{P}}(1).
    $}
\end{equation*}
\end{proposition}
This result simply claims that as the number of in-task samples $k$ increases, our meta CP will converge towards valid coverage for all tasks, not just on average. 
By itself, this is not particularly inspiring: after all, standard CP also becomes viable as $k\rightarrow \infty$. Rather, the key takeaway is that this desirable behavior is nicely preserved in our meta setup as well. 
In Figure~\ref{fig:quantile_res} we demonstrate that our $\pmeta_{1-\epsilon}$ indeed progresses towards task-conditional coverage as $k$ grows.

\subsection{Meta-learned approximate conformal prediction}
Recall that a key assumption in the theoretical results established in the previous section is that the distribution functions of our calibrations tasks, $F_i$ where $i \in \ical$, are known. In this section we turn to analyze the (much more common) setting where these $F_i$ must instead be estimated empirically. In this case, Theorem~\ref{thm:meta_cp} holds in expectation over the samples chosen for the calibration tasks. Furthermore, standard concentration results suggest that we can  approximate $F_i$ with little error, given enough empirical samples (which, in general, we assume we have for our  calibration tasks). We now further adapt Theorem~\ref{thm:meta_cp} to be conditionally valid with respect to the labeled examples that are used when replacing each task  $F_i$ with its plug-in estimate, $\widehat{F}_{m_i}$. 

First, we formalize a PAC-type 2-parameter validity definition (similar to \emph{training conditional} CP in \citet{vovk2012conditional}):

\label{sec:extensions}
\begin{definition}[$(\delta,\epsilon)$ task validity] $\mset$ is \emph{$(\delta,\epsilon)$ task valid} if for any task distribution $\ptask$, $\epsilon \in (0, 1)$, and $\delta \in (0, 1)$,
\begin{equation}
    \label{eq:delta_task_valid}
    \mathbb{P}\Big(\mathbb{P}\Big(Y^{\mathrm{test}}_{t+1} \in \mset\left(X^{\mathrm{test}}_{t+1}\right) \Big) \geq 1 - \epsilon\Big) \geq 1 - \delta.
\end{equation}
\end{definition}
\vspace{-5pt}
The outer probability is taken with respect to the data samples used for calibration. The basic idea here is to include a secondary confidence level $\delta$ that allows us to control how robust we are to sampling variance in our estimation of calibration tasks quantiles when computing $\Lambda(\beta; \ical)$, our conformal prediction correction factor.
We define a sample-conditional approach that is $(\delta, \epsilon)$ task valid, as follows:

\begin{proposition}[Sample-conditional meta CP]
\label{prop:sample_cp}
Assume that all $|\ical| = l$ calibration tasks are i.i.d., where for each task we have a fixed dataset that is also i.i.d. 
That is, for task $T_i$, we have drawn $m_i$ i.i.d. training examples, $\big(x_{i,j}, y_{i,j}\big)$, $j = 1,\ldots,m_i$. For any $\delta \in (0, 1)$, $\epsilon \in (0,1)$, and $\alpha \in \big(0, 1 - (1 - \delta)^{\frac{1}{n}}\big)$, define the adjusted $\epsilon'$ as
\begin{align}
\label{eq:correction}
{\epsilon}' &\leq \epsilon - \sqrt{\frac{-2}{l^2}\bigg(\sum_{i \in \ical} \gamma_i^2 \bigg)\log\bigg(1 - \frac{1 - \delta}{(1 - \alpha)^l}\bigg)}
\end{align}
where $\gamma_i =  \sqrt{\frac{\log(2/\alpha)}{2m_i}}$. Then $\mathcal{M}_{\epsilon'}(X_{t+1}^{\mathrm{test}})$ satisfies Eq.~\eqref{eq:delta_task_valid}.
\end{proposition}
\begin{remark}
We are free to choose $\alpha$ so as to optimize $\epsilon'$.
\end{remark}

Increasing the number of auxiliary tasks or samples per task make $\epsilon'$ closer to $\epsilon$.
In \S\ref{sec:results} we show that we can achieve tight prediction sets in practice, even  with small tolerances.
\section{Experimental Setup}
\label{sec:exp-set}

\begin{figure*}[!t]
\small
\centering
\footnotesize
\begin{subfigure}{0.32\textwidth}
\includegraphics[width=1.05\linewidth]{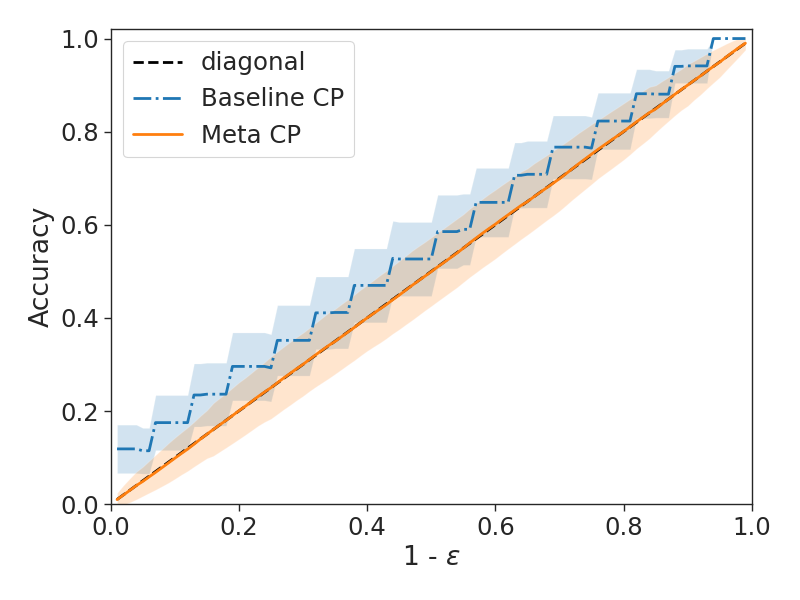} 
\end{subfigure}
~
\begin{subfigure}{0.32\textwidth}
\includegraphics[width=1.05\linewidth]{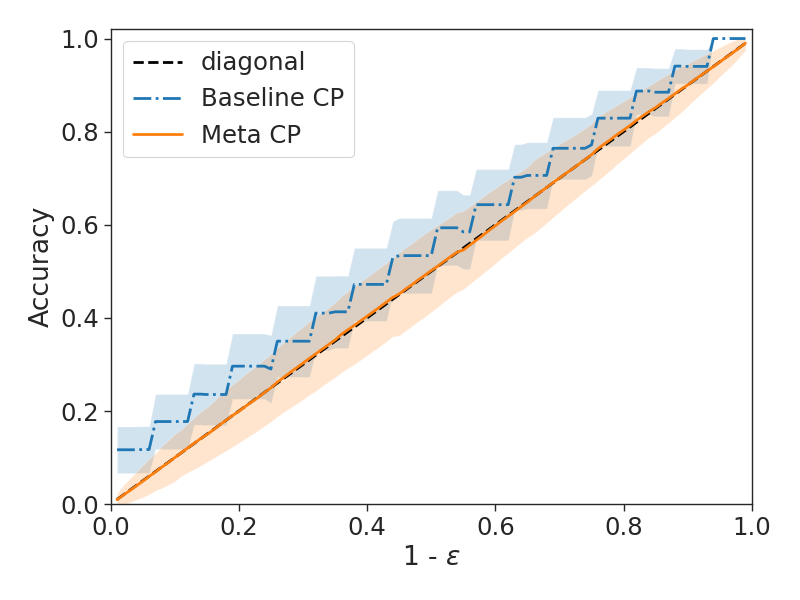}
\end{subfigure}
~
\begin{subfigure}{0.32\textwidth}
\includegraphics[width=1.05\linewidth]{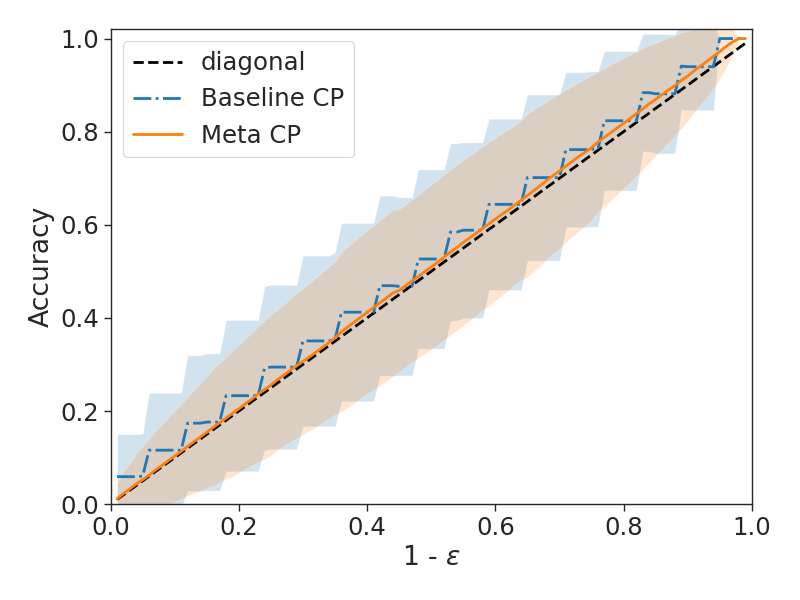} 
\end{subfigure}

\vspace*{-0.8\baselineskip}
\begin{subfigure}{0.32\textwidth}
\includegraphics[width=1.05\linewidth]{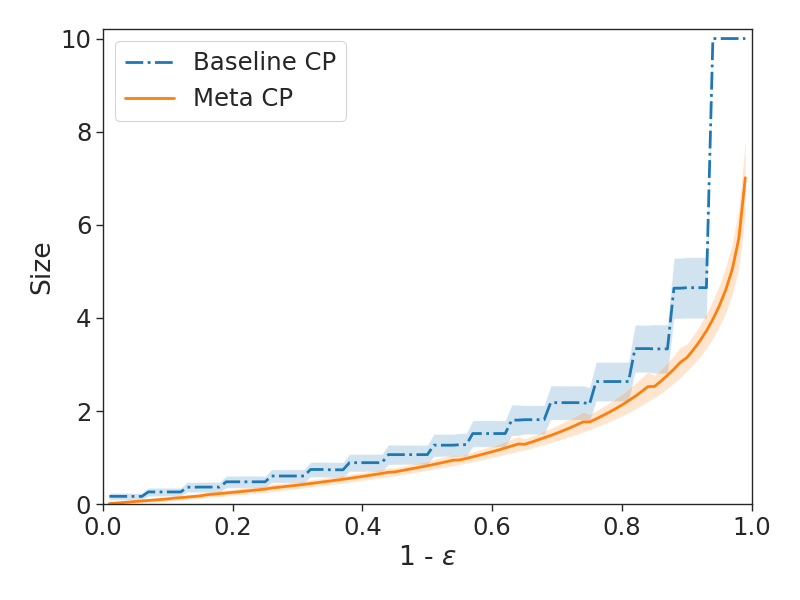} 
\caption{Image classification}
\end{subfigure}
~
\begin{subfigure}{0.32\textwidth}
\includegraphics[width=1.05\linewidth]{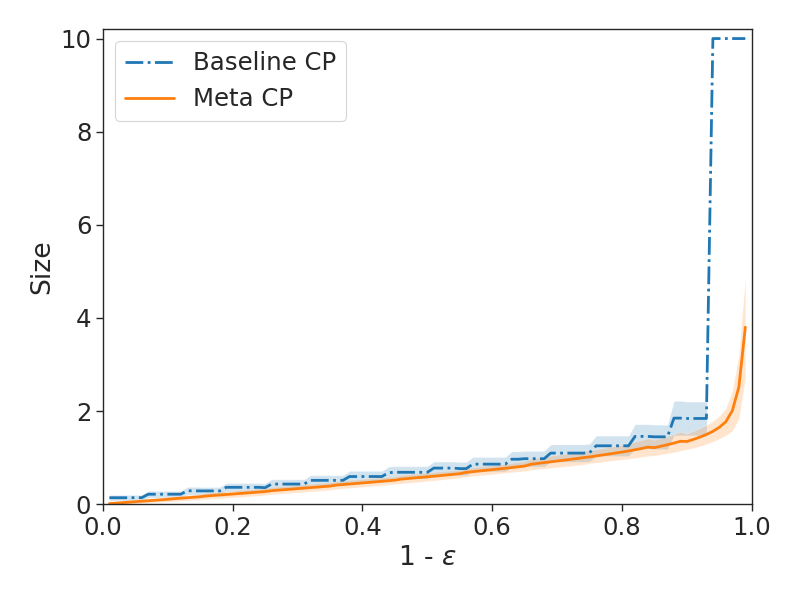}
\caption{Relation classification}
\end{subfigure}
~
\begin{subfigure}{0.32\textwidth}
\includegraphics[width=1.05\linewidth]{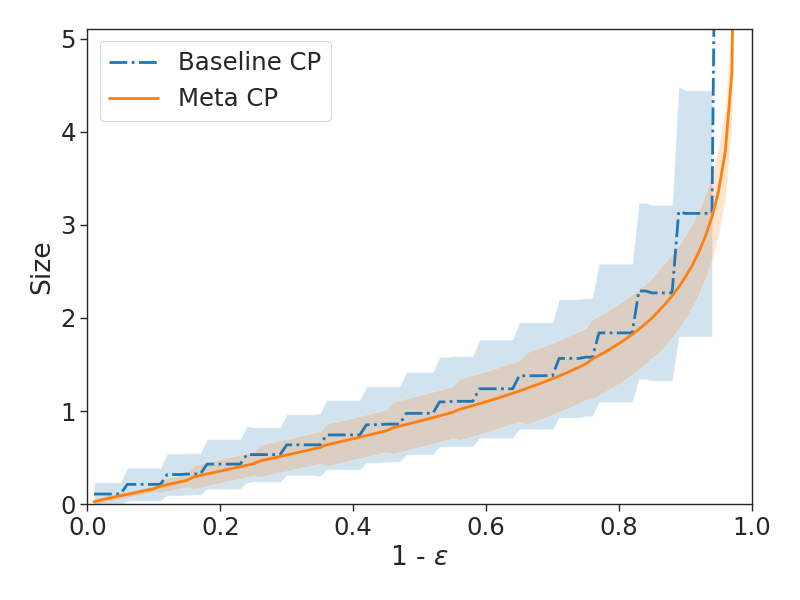} 
\caption{Chemical property prediction}
\end{subfigure}
\vspace*{-0.5\baselineskip}
\caption{Few-shot CP results as a function of $\epsilon$. The size of the prediction set of our meta CP approach is significantly better (i.e., \emph{smaller}) than that of our full CP baseline. Furthermore, our meta CP approach's average accuracy level is close to the diagonal---allowing it to remain valid in the sense of Eq.~\eqref{eq:task_valid}, but also less conservative when making predictions. Note that we care more about the right-hand-side behavior of the above graphs (i.e., larger $1-\epsilon$), as they correspond to higher coverage guarantees.}
\vspace{-5pt}
\label{fig:cp_res}
\end{figure*}



\subsection{Evaluation tasks}
\newpar{Image classification (CV)} As introduced in \S\ref{sec:introduction}, the goal of few-shot image classification is to train a computer vision model that generalizes to entirely new image classes at test time. We use the \emph{mini}ImageNet dataset~\citep{vinyals2016matching}, a downsampled version of a subset of classes from ImageNet~\cite{deng2009imagenet}. \emph{mini}ImageNet contains 100 classes that are divided into training, validation, and test class splits. Within each class partition, we construct $K$-shot $N$-way tasks, where $K$ examples per class are used to discriminate between a sample of $N$ distinct, novel classes. We use $K = 16$ and $N = 10$ in our experiments, for a total of $k = 160$ training examples. In order to avoid label imbalanced accuracy, however, we choose to focus on Mondrian CP~\cite{vovk2005algorithmic}, where validity is guaranteed across class type. Our meta nonconformity measure consists of a prototypical network on top of a CNN encoder.

\newpar{Relation classification (NLP)} Relation classification focuses on identifying the relationship between two entities mentioned in a given natural language sentence. In few-shot relation classification, the goal is to train an NLP model that generalizes to entirely new entity relationship types at test time. We use the FewRel 1.0 dataset~\cite{han2018fewrel}, which consists of 100 relations derived from 70$k$ Wikipedia sentences. Like \emph{mini}ImageNet, the relation types are divided into training, validation, and test splits.\footnote{We only use training/validation splits (the test set is hidden).} Within each partition, we sample $K$-shot $N$-way classification episodes (again with $K=16$ and $N = 10$ and Mondrian CP, as in our CV task). Our meta nonconformity measure consists of a prototypical network on top of a CNN encoder with GloVe embeddings~\cite{pennington2014glove}.

\newpar{Chemical property prediction (Chem)} In-silico screening of chemical compounds is an important task for drug discovery. Given a new molecule, the goal is to predict its activity for a target chemical property. We use the ChEMBL dataset~\cite{mayr}, and regress the pChEMBL value (a normalized log-activity metric) for individual molecule-property pairs. We select a subset of 296 assays from ChEMBL, and divide them into training (208), validation (44), and test (44) splits. Within each partition, each assay's pChEMBL values are treated as a regression task. We use $k = 16$ training samples per task. Our meta nonconformity measure consists of a few-shot, closed-form ridge regressor~\cite{bertinetto2018metalearning} on top of a directed Message Passing Network molecular encoder \cite{chemprop}.\footnote{We apply RRCM~\cite{nouretdinov2001rrcm} for full CP.}

\subsection{Evaluation metrics}
For each experiment, we use proper training, validation, and test meta-datasets of tasks. We use the meta-training tasks to learn all meta nonconformity measures $\smeta$ and meta quantile predictors $\pmeta$. We perform model selection for CP on the meta-validation tasks, and report final numbers on the meta-test tasks. For all methods, we report marginalized results over 5000 random trials, where in each trial we partition the data into $l$ calibration tasks ($T_{1:l}$) and one target task ($T_{t+1}$). In all plots, shaded regions show +/- the standard deviation across trials. We use the following metrics:

\newpar{Prediction accuracy} We measure accuracy as the rate at which the target label $y \in \mathcal{Y}$ is contained within the predicted label set. For classification problems, the prediction is a discrete set, whereas in regression the prediction is a continuous interval. To be valid, a conformal model should have an average accuracy rate $\geq 1 - \epsilon$. 

\newpar{Prediction size ($\mathbf{\shortdownarrow}$)} We measure the average size of the output (i.e., $|\cset|$) as a proxy for how \emph{precise} the model's predictions are. 
The goal is to make the prediction set as small as possible while still maintaining the desired accuracy. 

\subsection{Baselines}
For all experiments, we compare our methods to full conformal prediction, in which we use a meta-learned nonconformity scores---as defined in Eq.~\eqref{eq:meta_nonconf}. Though still a straightforward application of standard conformal calibration, meta-learning $\smeta$ with auxiliary tasks already adds significant statistical power to the model over an approach that would attempt to learn $\mathcal{S}$ from scratch for each new task.

In addition to evaluating improvement over full CP, we compare our approach to other viable heuristics for making set valued predictions: \texttt{Top-k} and \texttt{Naive}. In \texttt{Top-k} we always take the $k$-highest ranked predictions.  In \texttt{Naive} we select likely labels until the cumulative softmax probability exceeds $1 - \epsilon$. While seeming related to our CP approach, we emphasize that these are only heuristics, and do not give the same theoretical performance guarantees.

\section{Experimental Results}
\label{sec:results}
\begin{table*}[!t]
\centering
\small
\setlength{\tabcolsep}{10pt}
\begin{tabular}{cd|dd|dd|dd}
\toprule
Task &
  \multicolumn{1}{c}{Target Acc.} &
  \multicolumn{2}{c}{Baseline CP} &
  \multicolumn{2}{c}{Meta CP} &
  \multicolumn{2}{c}{$(\delta, \epsilon)$-valid Meta CP}\\
 &
  \multicolumn{1}{c}{$(1 - \epsilon)$} &
  \multicolumn{1}{c}{Acc.} &
  \multicolumn{1}{c}{$|\cset|$} &
  \multicolumn{1}{c}{Acc.} &
  \multicolumn{1}{c}{$|\mset|$} &
  \multicolumn{1}{c}{Acc.} &
  \multicolumn{1}{c}{$|\mathcal{M}_{k, \epsilon'}|$} \\
\midrule
\multirow{4}{*}{\textbf{CV}}   & 0.95 & 1.00 & 10.00 & 0.95 & 3.80 & 0.96 & 3.98 \\
                               & 0.90 & 0.94 & 4.22 & 0.90 & 2.85 & 0.91 & 2.96 \\
                               & 0.80 & 0.83 & 2.38 & 0.80 & 1.89 & 0.81 & 1.95 \\
                               & 0.70 & 0.76 & 1.94 & 0.70 & 1.37 & 0.71 & 1.42 \\
\midrule                               
\multirow{4}{*}{\textbf{NLP}} & 0.95 & 1.00 & 10.00 & 0.95 & 1.65 & 0.96 & 1.71\\
                            &  0.90 & 0.94 & 1.84 & 0.90 & 1.39 & 0.91&1.42\\
                            &  0.80 & 0.83 & 1.25 & 0.80 & 1.12 & 0.81& 1.14\\
                            &  0.70 & 0.76 & 1.10 & 0.70 & 0.93 & 0.71& 0.94\\
\midrule                               
\multirow{4}{*}{\textbf{Chem}} & 0.95 & 1.00 & \multicolumn{1}{r|}{$\mathrm{inf}$} & 0.97 & 3.44 & 0.99 & 5.25 \\
                               & 0.90 & 0.94 &  3.28 & 0.92 & 2.62 & 0.95 & 3.02 \\
                               & 0.80 & 0.82 &  2.08 & 0.82 & 1.95 & 0.86 & 2.16 \\
                               & 0.70 &  0.71& 1.59 & 0.72 & 1.56 & 0.76 & 1.70 \\
\bottomrule                               
\end{tabular}
\caption{Few-shot CP results for $\epsilon$ values. We report the empirical accuracy and raw prediction set size for our two meta CP methods, and compare to our baseline CP model (full CP with meta-learned $\smeta$). For our sample-conditional meta CP approach, we fix $\delta = 0.1$. Note that CP can produce empty sets if no labels are deemed conformal, hence the average classification size may fall below $1$ for high $\epsilon$.}
\vspace{-10pt}
\label{tab:main_results}
\end{table*}
In the following, we present our main conformal few-shot results. We evaluate both our sample-conditional and unconditional meta conformal prediction approaches. 

\newpar{Predictive efficiency} We start by testing how our meta CP approach affects the size of the prediction set. Smaller prediction set sizes correspond to more efficient conformal models. We plot prediction set size as a function of $\epsilon \in (0, 1)$ in Figure~\ref{fig:cp_res}. Table~\ref{tab:main_results} shows results for specific values of $\epsilon$, and also shows results for our \emph{sample-conditional} meta CP approach, where we fix $1- \delta$ at $0.9$ for all trials (note that the other meta results in Figure~\ref{fig:cp_res} and Table~\ref{tab:main_results} are \emph{unconditional}). Across all tasks and values of $\epsilon$, our meta CP performs the best in terms of efficiency. Moreover, the average size of the meta CP predictions increases smoothly as a function of $\epsilon$, while full CP suffers from discrete jumps in performance. Finally, we see that our sample-conditional $(\delta = 0.1, \epsilon)$ approach is only slightly more conservative than our unconditional meta CP method. This is especially true for domains with a higher number of auxiliary tasks and examples per auxiliary task (i.e., CV and NLP).

\newpar{Task validity} As per Theorem~\ref{thm:meta_cp}, we observe that our meta CP approach is valid, as the average accuracy always matches or exceeds the target performance level. Typically, meta CP is close to the target $1 - \epsilon$ level for all $\epsilon$, which indicates that it is not  overly conservative at any point (which improves the predictive efficiency). On the other hand, our full CP baseline is only close to the target accuracy when $1 - \epsilon$ is near a multiple of $\frac{1}{k+1}$. This is visible from its ``staircase''-like accuracy plot in Figure~\ref{fig:cp_res}. We see that our sample-conditional approach is slightly conservative, as its accuracy typically exceeds $1 - \epsilon$. This is more pronounced for domains with smaller amounts of auxiliary data.

\begin{figure}[!t]
    \centering
    \vspace{-5pt}
    \includegraphics[width=0.75\linewidth]{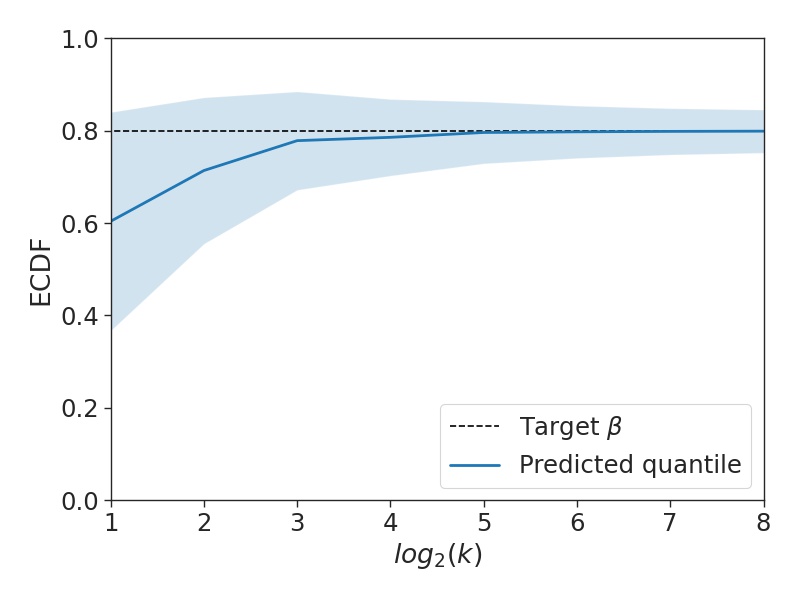} 
    \vspace{-10pt}
    \caption{We measure the error in our quantile predictor $\smeta_{\beta}$ (for $\beta = 0.8$) on the CV task as a function of $k$. As $k$ increases, the predictor begins to converge on an accurate $\beta$-quantile.} 
    \vspace{-12pt}
    \label{fig:quantile_res}
\end{figure}

\newpar{Conditional coverage} Figure~\ref{fig:quantile_res} shows the accuracy of our meta quantile predictor $\pmeta_{1-\epsilon}$ as a function of $k$. As expected, as $k$ grows, $\pmeta_{1 - \epsilon}$  becomes more accurate. This lessens the need for large correction factors $\Lambda(1- \epsilon, \ical)$, and leads to task-conditional coverage,  per Proposition~\ref{claim:asymptotic}.


\newcolumntype{?}{!{\vrule width 1.5pt}}
\begin{table}[!t]
\centering
\vspace{5pt}
\resizebox{\linewidth}{!}{%
\begin{tabular}{ccc?ccccc}
\toprule
\multicolumn{1}{l}{\textbf{\texttt{Top-k:}}}  & CV & NLP & \multicolumn{1}{l}{\textbf{\texttt{Naive:}}} & \multicolumn{2}{c}{CV} & \multicolumn{2}{c}{NLP} \\
Size ($k$) & Acc. & Acc. & Target Acc. & Acc. & Size & Acc. & Size \\
\midrule
5   & 0.96 & 0.99 & 0.95        & 0.97 & 4.38 & 0.99 & 2.98 \\
3   & 0.88 & 0.98 & 0.90        & 0.94 & 3.50 & 0.99 & 2.45 \\
1   & 0.60 & 0.79 & 0.80        & 0.88 & 2.61 & 0.97 & 1.94 \\ \bottomrule
\end{tabular}%
}
\caption{Non-conformal baseline heuristics (for classification tasks only). \texttt{Top-k} takes a target size ($k$), and yields statically sized outputs. \texttt{Naive} takes a target accuracy of $1 - \epsilon$, and yields dynamically sized outputs according to softmax probability mass.}
\label{tab:baselines}
\vspace{-10pt}
\end{table}

\newpar{Baseline comparisons} Table~\ref{tab:baselines} gives the results for our non-conformal heuristics, \texttt{Top-k} and \texttt{Naive}. We see that both approaches under-perform our CP method in terms of efficiency. Comparing to  Table~\ref{tab:main_results}, we see that we achieve similar accuracy to \texttt{Top-k} with smaller average sets (while also being able to set $\epsilon$). Similarly, \texttt{Naive} is uncalibrated and gives conservative results: for a target $\epsilon$ we obtain tighter prediction sets with our meta CP approach.

\section{Conclusion}
\label{sec:conclusion}
The ability to provide precise performance guarantees and make confidence-aware predictions is a critical element for many machine learning applications in the real world. Conformal prediction can afford remarkable finite-sample theoretical guarantees, but will suffer in practice when data is limited. In this paper, we introduced a novel and theoretically grounded approach to meta-learning few-shot conformal predictor using exchangeable collections of auxiliary tasks. Our results show that our method consistently improves performance across multiple diverse domains, and allow us to obtain meaningful and confident conformal predictors when using only a few in-task examples.

\section*{Acknowledgements}
We thank Kyle Swanson, the MIT NLP group, and anonymous reviewers for valuable feedback.
AF is supported in part by the NSF GRFP. TS is supported in part by DSO grant DSOCL18002. This work is also supported in part by MLPDS and the DARPA AMD project.

\bibliography{bib}
\bibliographystyle{icml2021}

\clearpage
\appendix
\counterwithin{figure}{section}
\counterwithin{table}{section}
\section{Proofs}
\label{app:proofs}
\begin{table*}[ht!]
\centering
\small
\begin{tabular}{cp{12cm}}
\toprule
Symbol &
  Meaning \\ \midrule
$k$ &
  The number of in-task examples used for few-shot learning. \\
$\epsilon$ &
  The stipulated performance tolerance. \\
$\delta$ &
  The stipulated secondary confidence tolerance for calibration conditional validity. \\
  \midrule
$\mathcal{T}$ &
  The space of potential tasks to be solved in a few-shot learning setting. \\
$\mathcal{X} \times \mathcal{Y}$ &
  The joint input ($X$'s) and output ($Y$'s) space. \\
$\mathcal{I}_{\mathrm{train}}$ &
  The set of auxiliary tasks used for meta-learning nonconformity scores and quantile predictors. \\
$\mathcal{I}_{\mathrm{cal}}$ &
  The set of auxiliary tasks used to calibrate the quantile predictor. \\ \midrule
$T_{t+1}$ &
  The target few-shot test task to be solved. \\
$\smeta$ &
  A meta-learned nonconformity measure. \\
$\pmeta_{1 - \epsilon}$ &
  A meta-learned regressor of the $1 - \epsilon$ quantile of $\smeta$'s scores on $T_{t+1}$ given $k$ in-task samples. \\
$\widehat{V}_{i, j}^{(x, y)}$ &
  The meta nonconformity score for example $j$ of task $i$, given the current candidate output $(x, y)$. \\
\midrule
$F_i, \widehat{F}_{m_i}$ &
  The true vs. $m_i$-sample empirical distribution function over nonconformity scores of task $i$.  \\
$\widehat{Q}_i$, $\mathrm{Quantile}(1 - \epsilon, F_i)$ &
  The predicted vs. true nonconformity score $1- \epsilon$ quantiles for task $i$. \\
$\Lambda(1 - \epsilon, \ical)$ &
  The $1 - \epsilon$ meta quantile correction factor, computed using calibration tasks. \\ 
  \midrule
  $\cset$, $\mset$ & Output label sets for standard and meta conformal prediction, respectively, at level $ 1 - \epsilon$. \\
\bottomrule
\end{tabular}%
\caption{Definitions of selected common notations used in this paper.}
\label{tab:symbols}
\vspace{-5pt}
\end{table*}
\subsection{Proof of Lemma~\ref{lemma:quantile}}
\begin{proof}
This is a well-known result; we prove it for completeness (see also \citet{tibshirani2019covariate} for an identical proof). Given support points $v_1, \ldots, v_n \in \mathbb{R}$ for a discrete distribution $F$, let $q = \mathrm{Quantile}(\beta; F)$. Any points $v_i > q$ do not affect this quantile, i.e., if we consider a new distribution $\tilde{F}$ where all points $v_i > q$ are mapped to arbitrary values also larger than $q$. then $\mathrm{Quantile}(\beta; F) = \mathrm{Quantile}(\beta; \tilde{F})$. Accordingly, for the nonconformity scores $V_i$, we have that
\begin{align*}
    V_{n+1} &> \mathrm{Quantile}(\beta; V_{1:n} \cup \{\infty\}) \Longleftrightarrow \\
    &V_{n+1} > \mathrm{Quantile}(\beta; V_{1:(n+1)}).
\end{align*}
Equivalently, we also have that
\begin{align*}
    V_{n+1} &\leq \mathrm{Quantile}(\beta; V_{1:n} \cup \{\infty\}) \Longleftrightarrow \\
    &V_{n+1} \leq \mathrm{Quantile}(\beta; V_{1:(n+1)}).
\end{align*}
Given the discrete distribution over the $n+1$ $V_i$, $V_{n+1} \leq \mathrm{Quantile}(\beta; V_{1:(n+1)})$ implies that $V_{n+1}$ is among the $\lceil\beta(n+1)\rceil$ smallest of $V_{1:(n+1)}$.
By exchangeability, this event occurs with probability at least $\frac{\lceil\beta(n+1)\rceil}{n+1} \geq \beta$. \end{proof}

\subsection{Proof of Theorem~\ref{thm:conformalprediction}}
\begin{proof}
This is also a well-known result; we prove it here for completeness (and see \citet{tibshirani2019covariate} for an identical proof). For notational convenience, let $V_i := V_i^{(X_{n+1}, Y_{n+1})}$. $Y_{n+1}$ is included in $\cset(X_{n+1})$ iff $V_{n+1} \leq \mathrm{Quantile}(1 - \epsilon; V_{1:n} \cup \{\infty\})$. As the nonconformity measure $\mathcal{S}$ preserves exchangeability by construction, if $(X_i, Y_i)$ for $i = 1,\ldots, n+1$ are exchangeable,  then so to are the nonconformity scores $V_i$, $i = 1,\ldots,n+1$. We can then apply Lemma~\ref{lemma:quantile} to complete the proof.
\end{proof}

\subsection{Proof of Lemma~\ref{lemma:meta_calibration}}
\begin{proof}
Let the event $\{T_i = t_i\}$ indicate that task $i$ has a quantile prediction $\{\widehat{Q}_i = q_i)$ and distribution function $\{F_i = f_i\}$ over meta nonconformity scores given $\smeta$.

For notational convenience, assume tasks $T_i$, $i \in \ical$ and $T_{t+1}$ are indexed contiguously as $i = 1,\ldots, n+1$. Next, denote by $E_t$ the event that $\{T_1, \ldots, T_{n+1}\} = \{t_1, \ldots, t_{n+1}\}$, i.e., we observe an \underline{unordered} set of task values. Exchangeability of tasks $T_i$ implies that
\begin{equation*}
    \mathbb{P}(T_{n+1} = t_i \mid E_t) = \frac{1}{n+1},
\end{equation*}
and, accordingly, that the distribution of $T_{n+1}\mid E_t$ is uniform on the set $\{t_1, \ldots, t_{n+1}\}$.

Again for notational convenience, let 
$$\widehat{V}_{i} := \widehat{V}_{i, k+1}^{(X_{i}^{\mathrm{test}}, Y_{i}^{\mathrm{test}})}$$
i.e., we use $\widehat{V}_{i}$ to denote the meta nonconformity score for task $i$'s random test point.

For any scalar $\lambda \in \mathbb{R}$, we can then write
\begin{align*}
    &\mathbb{P}(\widehat{V}_{n+1} \leq \widehat{Q}_{n+1} + \lambda \mid E_t) \\
    &= \sum_{i=1}^{n+1} \mathbb{P}(\widehat{V}_{n+1} \leq \widehat{Q}_{n+1} + \lambda, T_{n+1} = t_i \mid E_t) \\
    &= \sum_{i=1}^{n+1} \mathbb{P}(\widehat{V}_{n+1} \leq \widehat{Q}_{n+1} + \lambda \mid T_{n+1} = t_i) \mathbb{P}(T_{n+1} = t_i \mid E_t) \\
    &= \frac{1}{n+1}\sum_{i=1}^{n+1} \mathbb{P}(\widehat{V}_{n+1} \leq \widehat{Q}_{n+1} + \lambda \mid T_{n+1} = t_i).
\end{align*}
Since the event $\{T_{n+1} = t_i\}$ implies $\{\widehat{Q}_{n+1} = q_i, F_{n+1} = f_i\}$, we can  reduce this to
\begin{align*}
    \mathbb{P}(\widehat{V}_{n+1} \leq \widehat{Q}_{n+1} + \lambda \mid E_t)
    &= \frac{1}{n+1} \sum_{i=1}^{n+1} f_i(q_i + \lambda).
\end{align*}
Furthermore, on the event $E_t$, we have $\{T_1, \ldots, T_{n+1}\} = \{t_1, \ldots, t_{n+1}\}$, so  (with slight abuse of notation)
\begin{align*}
 \mathbb{P}(\widehat{V}_{n+1} \leq \widehat{Q}_{n+1} + \lambda \mid E_t)
     &= \frac{1}{n+1} \sum_{i=1}^{n+1} F_i(\widehat{Q}_i + \lambda).
\end{align*}
As $F_i$ is a distribution function with range $[0, 1]$, we can remove $T_{n+1}$ from the summation to get a lower bound,
\begin{align*}
\label{eq:dist}
 \mathbb{P}(\widehat{V}_{n+1} \leq \widehat{Q}_{n+1} + \lambda \mid E_t)
    &\geq \frac{1}{n+1} \sum_{i=1}^{n} F_i(\widehat{Q}_i + \lambda).
\end{align*}
For a fixed $\beta$, substitute $\Lambda(\beta;\ical)$ for $\lambda$ to derive
\begin{align*}
    \mathbb{P}(\widehat{V}_{n+1} &\leq \widehat{Q}_{n+1} + \Lambda(\beta;\ical) \mid E_t)
    \\ &\geq \frac{1}{n+1} \sum_{i=1}^{n} F_i(\widehat{Q}_i + \Lambda(\beta;\ical)) \geq \beta.
\end{align*}
Because this is true for any $E_t$, we can marginalize to obtain
\begin{align*}
    \mathbb{P}(&\widehat{V}_{n+1} \leq \widehat{Q}_{n+1} + \Lambda(\beta;\ical))  \\ &= \int_{E_t} \mathbb{P}(\widehat{V}_{n+1} \leq \widehat{Q}_{n+1} + \Lambda(\beta;\ical) \mid E_t)~d\mathbb{P}(E_t)\\
    &\geq \beta \int_{E_t} d\mathbb{P}(E_t)
    = \beta.
\end{align*}
\end{proof}

\subsection{Proof of Theorem~\ref{thm:meta_cp}}
\begin{proof} Again, for notational convenience, let
$$\widehat{V}_{i} := \widehat{V}_{i, k+1}^{(X_{i}^{\mathrm{test}}, Y_{i}^{\mathrm{test}})}$$
$Y_{t+1}^{\mathrm{test}}$ is included in $\mset(X_{t+1}^{\mathrm{test}})$ iff $\widehat{V}_{t+1} \leq \widehat{Q}_{t+1} + \Lambda(1 - \epsilon; \ical)$. As $\smeta$ and $\pmeta$ are trained on the disjoint proper training set $\itrain$, they preserve exchangeability, and produce exchangeable $\widehat{Q}_i$. We can then apply Lemma~\ref{lemma:meta_calibration}.
\end{proof}

\subsection{Proof of Proposition~\ref{claim:asymptotic}}
\begin{proof}
 Again, for notational convenience, let
$$\widehat{V}_{i} := \widehat{V}_{i, k+1}^{(X_{i}^{\mathrm{test}}, Y_{i}^{\mathrm{test}})}.$$
As stated in the claim, assume that as $k \rightarrow \infty$, 
\begin{equation*}
    \big|\pmeta_{1- \epsilon}(Z_{i,1:k};\phi_{\mathrm{meta}}) -\mathrm{Quantile}(1 - \epsilon, F_i)\big| = o_{\mathbb{P}}(1),
\end{equation*}
where $F_i$ is the distribution of $\widehat{V}_i$. That is, the quantile converges in probability to the true quantile where $\forall \alpha, \mu$  there exists $K_{\alpha, \mu}$ such that
\begin{equation*}
\resizebox{1\hsize}{!}{$\displaystyle
\mathbb{P}\left(\big|\pmeta_{1- \epsilon}(Z_{i,1:k};\phi_{\mathrm{meta}}) -\mathrm{Quantile}(1 - \epsilon, F_i)\big| \geq \mu \right)\leq \alpha,
$}
\end{equation*}
$\forall k > K_{\alpha, \mu}$. This is a standard property of consistent estimators (e.g., see \citet{lei2018distribution} for similar assumptions).

As $\Lambda(1-\epsilon; \ical) \geq 0$, for any target task $t_{t+1} \in \mathcal{T}$, we have that the corrected quantile, $\widetilde{Q}_{t+1}  = \widehat{Q}_{t+1} + \Lambda(1-\epsilon; \ical)$, is always \emph{conservative} for large enough $k$, i.e., 
\begin{equation}
\label{eq:convergence}
    \begin{split}
    \mathbf{1}\Big\{\widetilde{Q}_{t+1}  \geq \mathrm{Quantile}(1 - \epsilon, F_{t+1}) \mid T_{t+1} = t_{t+1}\Big\} \\
    = 1 - o_{\mathbb{P}}(1).
\end{split}
\end{equation}

In other words, this is to say that if $\pmeta_{1 - \epsilon}$ converges in probability to the true quantile, then $\pmeta_{1- \epsilon}$ $+$ some nonzero factor converges in probability to at least the true quantile.

 Next, if $\widehat{V}_{t+1} \leq \widetilde{Q}_{t+1}$, then $Y_{t+1}^{\mathrm{test}}$ is included in $\mset(X_{t+1}^{\mathrm{test}})$ (according to the definition of $\mset$).  Furthermore, by the definition of $\mathrm{Quantile}$, the event that $\widehat{V}_{t+1} \leq \mathrm{Quantile}(1 - \epsilon, F_{t+1})$ happens with probability at least $1 - \epsilon$. 
 Therefore, if $\widetilde{Q}_{t+1}  \geq\mathrm{Quantile}(1 - \epsilon, F_{t+1})$, then $Y_{t+1}^{\mathrm{test}}$ is included in $\mset(X_{t+1}^{\mathrm{test}})$ with probability at least $1 - \epsilon$.
Combining  with Eq.~\eqref{eq:convergence} completes our proof.
\end{proof}

\subsection{Proof of Proposition~\ref{prop:sample_cp}}
\begin{proof}
Let $\widehat{F}_{m_i}$ be the $m_i$-sample ECDF for $T_i$. Define the empirical correction, $\Lambda'(\beta, \ical)$, when plugging in $\widehat{F}_{m_i}$  as
\begin{equation}
\label{eq:lambda_prime}
\resizebox{.88\hsize}{!}{$\displaystyle
  \inf\bigg\{\lambda \colon \frac{1}{|\ical|+1} \sum_{i \in \ical} \widehat{F}_{m_i}\big(\vtest \leq \widehat{Q}_i + \lambda\big) \geq \beta\bigg\}
    $}
\end{equation}
where the ECDF is calculated as
\begin{equation*}
    \widehat{F}_{m_i} := \sum_{j=1}^{m_i}\mathbf{1}\{ \widehat{V}_{i,k+1}^{(j)} \leq \widehat{Q}_i + \lambda\},
\end{equation*}
where $\widehat{V}_{i,k+1}^{(j)}$ are i.i.d. and  $\widehat{V}_{i,k+1}^{(j)} \overset{d}{=} \widehat{V}_{i,k+1}^{\mathrm{test}}$.

We proceed in two parts. First, we prove that if the approximation error incurred by using $\Lambda'(\beta, \ical)$ is bounded by $\tau$ with probability $1 - \delta$, then $\mathcal{M}_{\epsilon - \tau}$ is $(\delta, \epsilon)$ valid. Second, we prove that the error is bounded according to Eq.~\eqref{eq:correction}.

(1) 
Following the proof of Lemma~\ref{lemma:meta_calibration}, we have that
\begin{align}
\label{eq:approx}
\begin{split}
    \mathbb{P}(\widehat{V}_{t+1} &\leq \widehat{Q}_{t+1} + \Lambda'(\beta;\ical) \mid E_t)
    \\ &\geq \frac{1}{|\ical|+1} \sum_{i \in \ical} F_i(\widehat{Q}_i + \Lambda'(\beta;\ical)).
\end{split}
\end{align}

For ease of notation, let
\begin{align*}
    A &:=\frac{1}{|\ical|+1} \sum_{i \in \ical} \widehat{F}_{m_i}(\widehat{Q}_i + \Lambda'(\beta;\ical)), \\
    B &:= \frac{1}{|\ical|+1} \sum_{i \in \ical} F_i(\widehat{Q}_i + \Lambda'(\beta;\ical)).
\end{align*}

Next, assume that (to be proved) for some $\tau > 0$
\begin{equation}
\label{eq:bound_assumption}
\mathbb{P}(A - B < \tau) \geq 1 - \delta.
\end{equation}
By construction---see Eq.~\eqref{eq:lambda_prime}---we have $A \geq \beta$. Then by Eq.~\eqref{eq:bound_assumption}, we have that with probability $1 - \delta$,
\begin{align*}
   B > A - \tau \geq \beta - \tau. 
\end{align*}
Choose $\beta \geq 1 - \epsilon + \tau$. Then $B \geq 1 - \epsilon$. By convention, this corresponds to $\beta := 1 - \epsilon' \geq 1 - (\epsilon - \tau)$, or $\epsilon' \leq \epsilon - \tau$ as in Eq.~\eqref{eq:correction}.
Combining this with Eq.~\eqref{eq:approx}, we have 
$$\mathbb{P}(\widehat{V}_{t+1} \leq \widehat{Q}_{t+1} + \Lambda'(\beta;\ical) \mid E_t) \geq 1 - \epsilon.$$
This is true for all $E_t$, so we can marginalize to obtain
$$\mathbb{P}(\widehat{V}_{t+1} \leq \widehat{Q}_{t+1} + \Lambda'(\beta;\ical)) \geq 1 - \epsilon.$$

(2) We now prove the assumption stated in Eq.~\eqref{eq:bound_assumption}. Given an $m$-sample ECDF, $\widehat{F}_{m}(u)$, for some random variable $U$, the Dvoretsky-Kiefer-Wolfowitz inequality allows us to build a confidence interval for the value of the true distribution function, $F(u)$, where
\begin{equation*}
\mathbb{P}\left( \sup_{u \in \mathbb{R}} |\widehat{F}_{m}(u) - F(u)| > \gamma\right) \leq 2e^{-2n\gamma^2}.
\end{equation*}
Alternatively stated, with probability at least $1 - \alpha$, $F(u) \in [\widehat{F}_{m}(u) - \gamma,  \widehat{F}_{m}(u) + \gamma]$, where $\gamma = \sqrt{\frac{\log\frac{2}{\alpha}}{2n}}$.

We combine this result with Hoeffding's inequality.

Let $Y_i := \widehat{F}_{m_i}(\widehat{V}_i \leq \widehat{Q}_i + \lambda) - F_i(\widehat{V}_i \leq \widehat{Q}_i + \lambda)$. Once again for notational convenience, assume tasks $T_i$, $i \in \ical$ and $T_{t+1}$ are indexed contiguously as $i = 1,\ldots, n+1$. The difference, $A - B$, is then equivalent to $\frac{1}{n+1} \sum_{i=1}^n Y_i$. According to our assumptions, $Y_i$'s are i.i.d., $\mathbb{E}[Y_i] = \mathbb{E}[\widehat{F}_{m_i}] - \mathbb{E}[F_i] = 0$, and $Y_i \in [-\gamma_i, \gamma_i]$ w.p. $1 - \alpha$. As above, we define $\gamma_i = \sqrt{\frac{\log\frac{2}{\alpha}}{2m_i}}$.

Applying Hoeffding's inequality gives
\begin{align*}
    &\mathbb{P}\Big(\frac{1}{n+1}\sum_{i=1}^n Y_i < \tau\Big) \\
    &\geq \mathbb{P}\Big(\sum_{i=1}^n Y_i < n\tau , \bigcap_{i=1}^n Y_i \in  [-\gamma_i, \gamma_i]\Big) \\
    &\geq \mathbb{P}\Big(\sum_{i=1}^n Y_i < n\tau \Big| \bigcap_{i=1}^n Y_i \in  [-\gamma_i, \gamma_i]\Big) \mathbb{P}\Big(\bigcap_{i=1}^nY_i \in  [-\gamma_i, \gamma_i]\Big)\\
    &\geq \Big(1 - e^{-\frac{2n^2\tau^2}{\sum_{i=1}^n(2\gamma_i)^2}}\Big) \Big(1 - \alpha\Big)^n.
\end{align*}
Solving for $\tau$ given the target $1 - \delta$ error probability yields
\begin{align*}
    1 - \delta &= \Big(1 - e^{-\frac{2n^2\tau^2}{\sum_{i=1}^n(2\gamma_i)^2}}\Big) \Big(1 - \alpha\Big)^n \\
    \tau &= \sqrt{\frac{-2}{n^2}\Big(\sum_{i=1}^{n} \gamma_i^2 \Big)\log\Big(1 - \frac{1 - \delta}{(1 - \alpha)^n}\Big)}
\end{align*}
This is valid for any choice of $\alpha$ (as long as the $\log$ term is defined), so we are free to choose $\alpha$ that minimizes $\tau$.
\end{proof}
\section{Meta Conformal Prediction Details}
\label{app:meta_cp_extra}

\subsection{Meta-Learning algorithms}
\label{app:meta_algs}

\newpar{Prototypical networks~\cite{snell2017prototypical}} We use prototypical networks for our classification tasks. We assume that for each task we have $N$ total classes with $K$ examples per class (for a total of $k = N \times K$ training examples). In this model, an encoder, $\mathbf{h} = \mathrm{enc}(x; \theta)$ is trained to produce vector representations. Thereafter, a ``prototype'' for each class is computed by averaging the representations of all instances of that class. Let $S_j$ denote the support set of training examples for class $j$. Then the prototype $\mathbf{c}_j$ is 
\begin{equation*}
    \mathbf{c}_j := \frac{1}{|S_j|}\sum_{(x_i, y_i) \in S_j} \mathrm{enc}(x_i; \theta).
\end{equation*}
The likelihood of each class is then calculated using a softmax over the euclidean distance to each prototype:
\begin{equation}
    \label{eq:proto}
    p_{\theta}(y = j \mid x) := \frac{\exp(-d(\mathbf{c}_j, \mathrm{enc}(x; \theta)))}{\sum_{j'} \exp(-d(\mathbf{c}_{j'}, \mathrm{enc}(x; \theta)))},
\end{equation}
where $d(\cdot, \cdot)$ denotes the euclidean distance.

During training, random training ``episodes'' are created by sampling $N$ classes from the training set. For each class, $K$ examples are randomly sampled to construct the prototypes. An additional $Q$ examples are then sampled to simulate queries. The optimization objective is to then minimize the cross entropy loss across queries.

After training, we use $-p_{\theta}(y = j \mid x)$ as defined in Eq.~\eqref{eq:proto} as the nonconformity measure for label $y = j$.



\newpar{Differentiable ridge regression~\cite{bertinetto2018metalearning}}
We use differentiable ridge regression networks for our regression tasks. We assume that for each task we have $k$ labeled $(x_i, y_i)$ pairs, where $y \in \mathbb{R}$. In this model, like the prototypical networks, an encoder, $\mathbf{h} = \mathrm{enc}(x; \theta)$, is trained to produce vector representations of dimension $d$. We then solve a least-squares regression  to obtain our prediction, $\hat{y} = \mathbf{w} \cdot \mathrm{enc}(x; \theta)$, where
\begin{equation*}
    \mathbf{w} = X^\top ( XX^\top + \lambda I)^{-1}Y
\end{equation*}
with $X \in \mathbb{R}^{k \times d}$, $Y \in \mathbb{R}^k$, and $\lambda$ a meta regularization parameter that we optimize. We optimize MSE by back-propagating through the least-squares operator to the encoder. We train using the same episode-based procedure that we described for the prototypical networks.

After training, we use the absolute error, $|\hat{y} - y|$, as the nonconformity score for candidate $y \in \mathbb{R}$.

\newpar{Deep sets~\cite{zaheer2017sets}} We use a simple deep sets architecture for all of our quantile predictors. Deep sets are of the form
\begin{equation*}
    f(X) := \mathrm{dec}\Big(\sum_{x \in X} \mathrm{enc}(x; \phi_1); \phi_2\Big)
\end{equation*}
where $X$ is an input set of elements, $\mathrm{enc}$ is an element-wise  encoder, and $\mathrm{dec}$ is a decoder that operates on the aggregated encoded set elements. Importantly, the deep sets model $f$ is invariant to permutations of the elements in $X$.

\subsection{Implementation details}

\newpar{Image classification}
Each image is first resized to $84 \times 84$ pixels. We use a CNN encoder with 4 layers. Each layer contains a $3 \times 3$ convolution kernel with $64$ channels and a padding of size 1, followed by batch normalization layer, ReLU activation, and a $2 \times 2$ max pooling filter. The final output is of size $1600$, which we use to compute the prototypes and as the query representations. We train the model for $100$ epochs with an Adam optimizer and a batch size of 256. In each epoch, we run $100$ episodes in which we sample $10$ support images and $15$ query images per class.

\newpar{Relation classification}
We use GloVe~\citep{pennington2014glove} word embeddings of size $50$ to convert the sentence into vectors. To each word embedding, we also concatenate two learned position embeddings of size $5$, where the positions are relative to the location of the two entities in the sentence. Thereafter, a 1D convolution is applied with $230$ output channels, a kernel size of 3 and padding size 1, followed by a ReLU activation. Finally, a max pooling filter is applied. The resultant sentence representation of size $230$ is used to compute the prototypes and query representations. We train the model for a total of $20k$ episodes with a SGD optimizer and a batch size of 32. In each episode, we sample $10$ support sentences and $5$ query sentences per class.

\newpar{Chemical property prediction} Our ridge regression network uses directed message passing networks~\cite{chemprop} to compute $\mathrm{enc}(x; \theta)$. The message passing network uses graph convolutions to learn a deep molecular representation that is shared across property predictions. We also include additional RDKit features as inputs.\footnote{\url{www.rdkit.org}} We map inputs with a FFNN with hidden size $200$, and then apply 3 layers of graph convolutions with a hidden size of $256$. Finally, we map the output representation to a hidden size of $16$, and apply least-squares regression. We train the network using an Adam optimizer for $15$ epochs with $8$ meta episodes per batch, each with $32$ queries (for a total batch size of $256$).

\newpar{Quantile prediction} For all of our quantile predictors, we use a 2-layer FFNN for both the element-wise encoder, $\mathrm{enc}(\cdot; \phi_1)$, and the aggregated set decoder, $\mathrm{dec}(\cdot; \phi_2)$. Each FFNN has a hidden size of $256$ and uses ReLU activations. We train the network using an Adam optimizer for $15$ epochs with batch size $64$.

\subsection{Training strategy}
\label{app:implementation}
\begin{figure}[!tbh]
    \centering
    \small
    \includegraphics[width=1\linewidth]{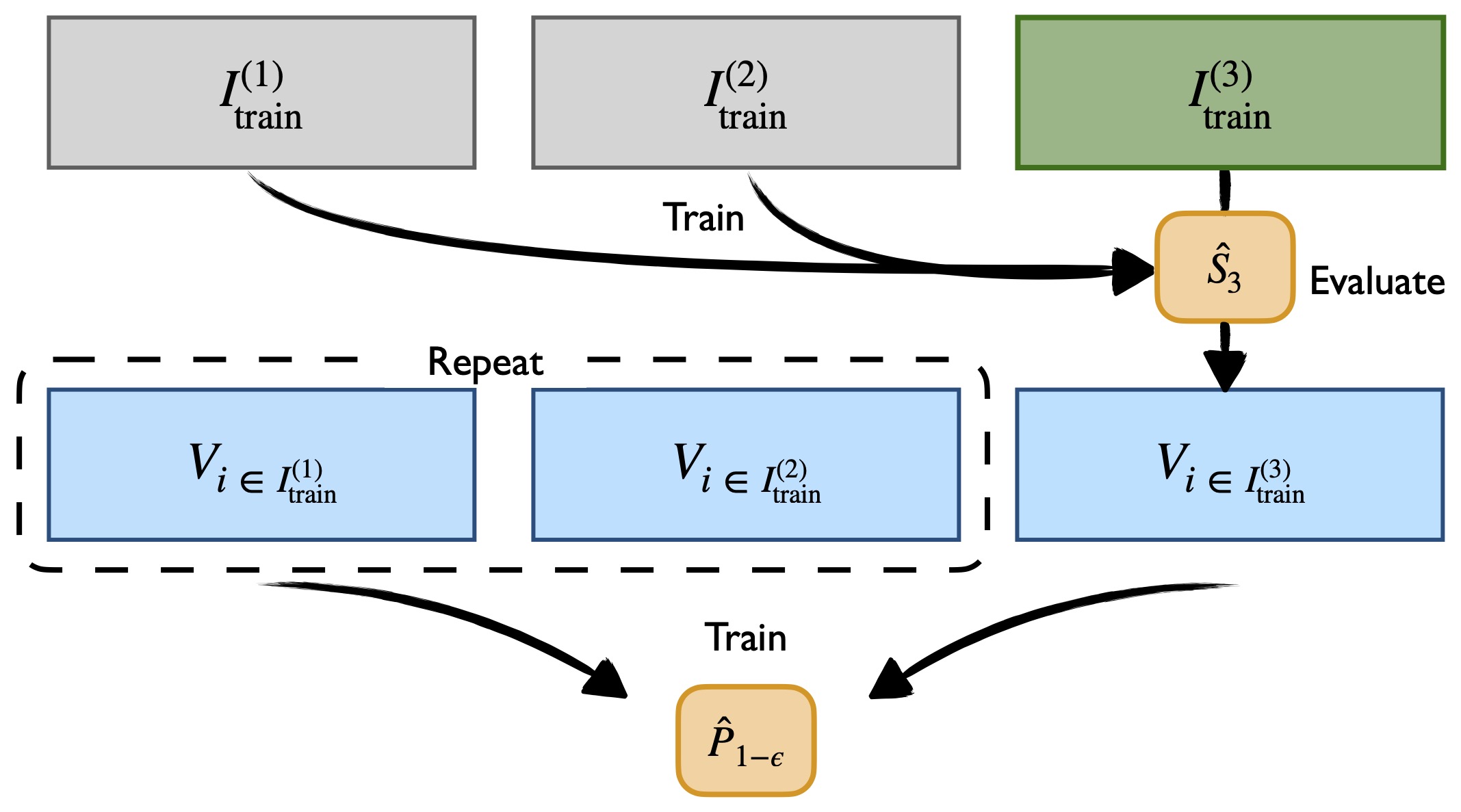}
    \caption{An illustration of our strategy for learning meta nonconformity measures, $\smeta$, and meta quantile predictors, $\pmeta_{1 - \epsilon}$. As $\pmeta_{1 - \epsilon}$ is trained on the outputs of $\smeta$, we adopt a cross-fold procedure where we first train $\smeta$ on a fraction of the data, and evaluate nonconformity scores on the held-out fold. We repeat this process for all $k_f$ folds, and then aggregate them all for training $\pmeta_{1 - \epsilon}$.}
    \label{fig:training}
\end{figure}

We adopt a cross-fold procedure for training our meta nonconformity measure $\smeta$ and meta quantile predictor $\pmeta_{1- \epsilon}$ in a data efficient way, as outlined in \S\ref{sec:metalearning}. Figure~\ref{fig:training} illustrates this cross-fold process, in which we train a meta-nonconformity measure on each training fold and aggregate their predictions as input data for the quantile predictor. 

Since we train in a cross-fold manner but ultimately use a meta-nonconformity measure $\smeta$ that is trained on \emph{all} of the training data, there is a train-test mismatch in the data supplied to the quantile predictor. Nevertheless, any error induced by this discrepancy (and any other sources of error, for that matter) is handled during meta-calibration (\S\ref{sec:metacalibration}).

All experiments took 1-5 hours to run on an Nvidia 2080 Ti GPU. As absolute performance is not the primary goal of this work, little hyperparameter tuning was done (most hyperparameters were taken from prior work). Datasets are available for \emph{mini}ImageNet\footnote{\url{https://github.com/yaoyao-liu/mini-imagenet-tools}}, FewRel 1.0\footnote{\url{https://thunlp.github.io/1/fewrel1.html}}, and ChEMBL\footnote{\url{https://github.com/chemprop/chemprop}}.

\end{document}